%% file: main.tex
\documentclass[10pt,twocolumn,letterpaper]{article}

\usepackage{iccv}
\usepackage{times}
\usepackage{epsfig}
\usepackage{graphicx}
\usepackage{amsmath}
\usepackage{amssymb}
\usepackage{booktabs}
\usepackage[table]{xcolor}
\usepackage{xspace}
\usepackage{nicefrac}
\usepackage{float}
\usepackage{sidecap,tabularx}
\usepackage[accsupp]{axessibility}

\definecolor{best}{RGB}{255, 220, 200}
\definecolor{second}{RGB}{255, 255, 200}

\usepackage{physics}
\usepackage{siunitx}

\newcommand{\thename}{EverLight\xspace}

\renewcommand{\etal}{et al.\xspace}

\renewcommand{\eg}{e.g.\xspace}

\usepackage[pagebackref=true,breaklinks=true,letterpaper=true,colorlinks,bookmarks=false]{hyperref}

\DeclareMathOperator*{\argmin}{arg\,min}

\usepackage[capitalize]{cleveref}
\crefname{section}{sec.}{secs.}
\Crefname{section}{Sec.}{Secs.}
\crefname{table}{tab.}{tabs.}
\Crefname{table}{Tab.}{Tabs.}
\crefname{figure}{fig.}{figs.}
\Crefname{figure}{Fig.}{Figs.}
\crefname{equation}{eq.}{eqs.}
\Crefname{equation}{Eq.}{Eqs.}

\iccvfinalcopy % *** Uncomment this line for the final submission

\ificcvfinal\fi

\begin{document}

%%%%%%%%% TITLE
\title{\thename: Indoor-Outdoor Editable HDR Lighting Estimation}

\author{Mohammad Reza Karimi Dastjerdi$^{1}$\thanks{Research partly done when Mohammad Reza was an intern at Adobe.},
Jonathan Eisenmann$^{2}$, Yannick Hold-Geoffroy$^{2}$,\\
Jean-Fran\c{c}ois Lalonde$^{1}$\\
$^1$Université Laval, $^2$Adobe\\
\small{\texttt{\url{https://lvsn.github.io/everlight/}}}
} 

\makeatletter
\g@addto@macro\@maketitle{
\vspace{-0.5in}
    \begin{figure}[H]
    \setlength{\linewidth}{\textwidth}
    \setlength{\hsize}{\textwidth}
    \centering
    \includegraphics[width=\linewidth]{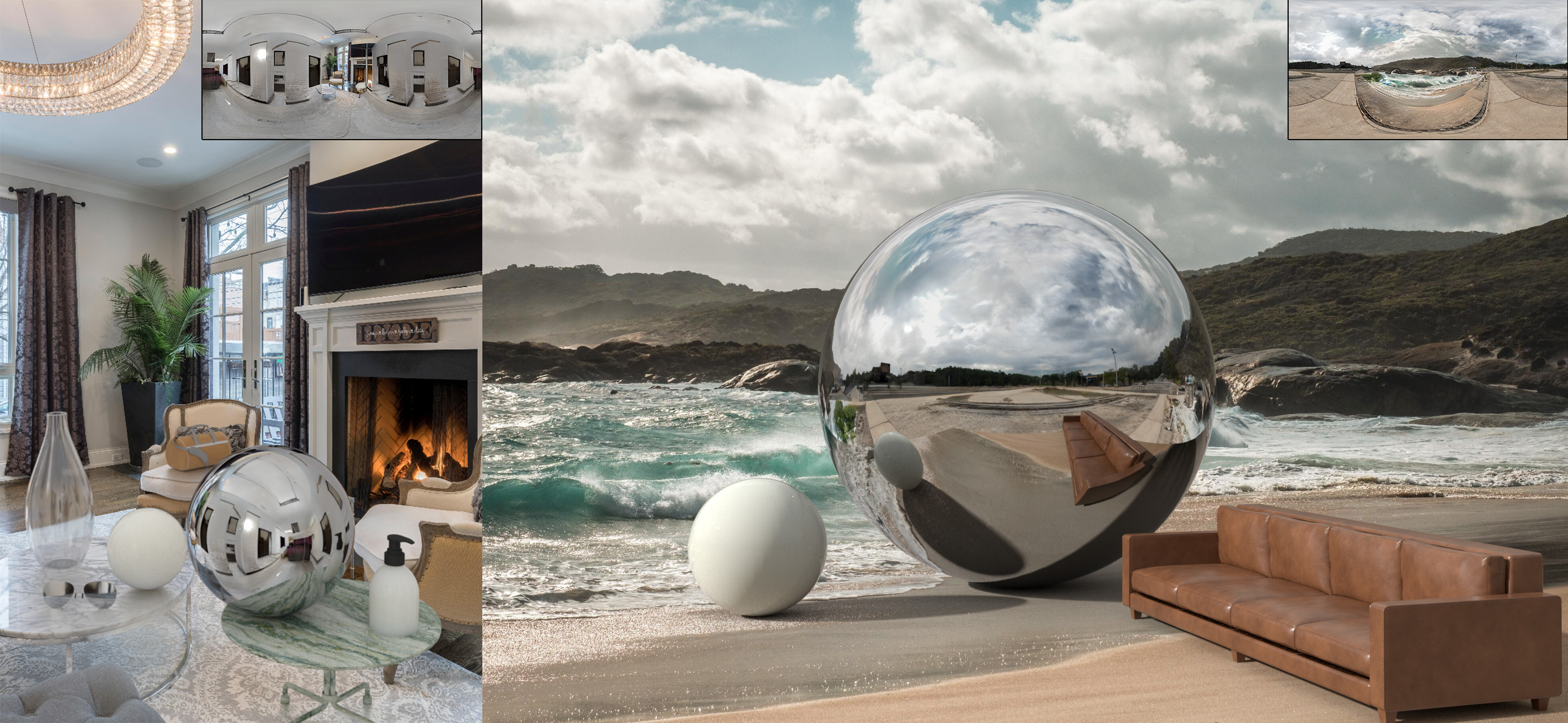}
    \caption{We present a method that produces an HDR \ang{360} environment map (top right insets) from regular images (background) captured indoors (left) or outdoors (right) with any camera. Our method increases the realism of reflections in specular materials such as metals and plastics. Our approach simultaneously estimates the light sources in the environment beyond the camera frame and models them as HDR spherical gaussians, ready for use as HDRI in a renderer. This increases realism without user interaction when inserting virtual objects (left: vase, glasses, spheres, handbag, hand sanitizer, right table; right: spheres, sofa). Our parametric lighting representation makes it easy for users to add, remove, and edit lights in the environment, which will plausibly react to those edits.}
    \label{fig:teaser}
    \end{figure}
}
\makeatother
\maketitle

\ificcvfinal\thispagestyle{empty}\fi

\begin{abstract}
% We propose a method for estimating lighting as editable HDR \ang{360} panoramas from regular images, regardless if they were taken indoors or outdoors. 
Because of the diversity in lighting environments, existing illumination estimation techniques have been designed explicitly on indoor or outdoor environments. Methods have focused specifically on capturing accurate energy (e.g., through parametric lighting models), which emphasizes shading and strong cast shadows; or producing plausible texture (e.g., with GANs), which prioritizes plausible reflections. Approaches which provide editable lighting capabilities have been proposed, but these tend to be with simplified lighting models, offering limited realism. 
In this work, we propose to bridge the gap between these recent trends in the literature, and propose a method which combines a parametric light model with \ang{360} panoramas, ready to use as HDRI in rendering engines. We leverage recent advances in GAN-based LDR panorama extrapolation from a regular image, which we extend to HDR using parametric spherical gaussians. To achieve this, we introduce a novel lighting co-modulation method that injects lighting-related features throughout the generator, tightly coupling the original or edited scene illumination within the panorama generation process. In our representation, users can easily edit light direction, intensity, number, etc. to impact shading while providing rich, complex reflections while seamlessly blending with the edits. Furthermore, our method encompasses indoor and outdoor environments, demonstrating state-of-the-art results even when compared to domain-specific methods. 
\end{abstract}

\input{1_intro}
\input{2_related_work}

\input{3_method}
\input{4_experiment}
\input{5_discussion}

\paragraph{Acknowledgements}
The name of this paper \emph{EverLight} is a homage to the Critical Role's \emph{The Legend of Vox Machina}. This work was partially supported by NSERC grant ALLRP 557208-20. We thank Sai Bi for his help with extending the dynamic range of the panoramas and everyone at UL who helped with proofreading.

{\small
\bibliographystyle{ieee_fullname}
\bibliography{bibliography}
}

%-------------------------------------------------------------------------
%-------------------------------------------------------------------------
%%%%%%%%% APPENDIX

% NONE

%-------------------------------------------------------------------------
%-------------------------------------------------------------------------
%-------------------------------------------------------------------------
\end{document}

%%%%%%%%% TITLE
\title{\LaTeX\ Author Guidelines for ICCV Proceedings}

\author{First Author\\
Institution1\\
Institution1 address\\
{\tt\small firstauthor@i1.org}
\and
Second Author\\
Institution2\\
First line of institution2 address\\
{\tt\small secondauthor@i2.org}
}

\maketitle
\ificcvfinal\thispagestyle{empty}\fi

\begin{itemize}
    \item Evaluation of the light prediction network? 
\end{itemize}

{\small
\bibliographystyle{ieee_fullname}
\bibliography{bibliography}
}

%-------------------------------------------------------------------------
%-------------------------------------------------------------------------
%%%%%%%%% APPENDIX

% NONE

%-------------------------------------------------------------------------
%-------------------------------------------------------------------------
%-------------------------------------------------------------------------

%% file: 1_intro.tex
%!TEX root = main.tex
\section{Introduction}

The realistic blending of virtual assets in real imagery is required in many scenarios, ranging from special effects to augmented reality (AR) and advanced image editing. In this context, ``getting the lighting right'' is one of the key challenges. Image-based lighting~\cite{debevec1998rendering} can be employed to solve this problem, but it requires physical access to the scene and specialized equipment. In an attempt to automate this process, techniques that learn to predict lighting directly from captured imagery have been proposed. While earlier approaches relied on engineered features~\cite{lalonde2012estimating}, they have since been replaced by learning-based techniques~\cite{hold2017deep,gardner2017learning}. 
Driven by the popularity of on-device AR applications, this line of research has recently attracted much attention, and several trends have emerged in the literature. 

Perhaps the most popular trend, naturally, has been to develop \emph{richer, more expressive lighting representations} to improve the accuracy of lighting estimations. The seemingly most popular representation is environment maps (equirectangular images representing the entire field of view in \ang{360})~\cite{gardner2017learning,song2019neural,legendre2019deeplight,holdgeoffroy2019deep,somanath2021hdr,wang2022stylelight}, but others have been explored as well, including spatially-varying spherical harmonics~\cite{garon2019fast}, parametric light sources~\cite{hold2017deep,zhang2019all,gardner2019deep}, dense spherical Gaussians in 2D~\cite{li2020inverse} or 3D~\cite{wang2021learning}, sparse spherical gaussians~\cite{zhan2021emlight,zhan2021gmlight}, sparse needlets~\cite{zhan2021sparse}, multi-scale volume of implicit features~\cite{srinivasan2020lighthouse}, and full neural light fields~\cite{wang2022neural}. Recently, hybrid approaches combining environment maps and a single parametric light have also been proposed~\cite{weber2022editable}. 

Another identifiable trend has been to create \emph{domain-specific approaches} to design the representation and/or approach specifically for a target domain. The most common way of defining a domain has been to explicitly consider indoor (\eg, \cite{gardner2017learning,garon2019fast,li2020inverse,wang2021learning}) vs outdoor (\eg, \cite{lalonde2012estimating,hold2017deep,zhang2019all,holdgeoffroy2019deep,zhu2021spatially}) domains. Of note, Legendre~\etal~\cite{legendre2019deeplight} proposed what is perhaps the only method in the literature that works for both indoor and outdoor scenes. 

A third, more recent trend are \emph{user-editable} methods, where the goal is to employ or design a lighting representation that can easily be understood and modified by a user. For example, parametric models~\cite{hold2017deep,zhang2019all,gardner2019deep} represent the dominant light sources using intuitive parameters (\eg, position, intensity, etc.) that can be interacted with easily, but fail to generate realistic reflections. Methods based on hybrid models~\cite{weber2022editable} or GAN inversion~\cite{wang2022stylelight} have demonstrated promising results, but are either limited to a single light source~\cite{weber2022editable} or employ a slow optimization process~\cite{wang2022stylelight}. 

% Finally, another closely-related trend does not explicitly perform lighting estimation, but aims to recover the \emph{full \ang{360} field of view} from the image~\cite{song2018im2pano3d,akimoto2019360,han2021piinet,hara2021spherical,karimi2022guided}. By leveraging large-scale datasets of \ang{360} panoramas (\eg, \cite{cruz2021zillow}) and powerful GAN architectures~\cite{zhao2021comodgan}, these techniques create high quality results across a large variety of scenes. However, these methods do not recover the high dynamic range light sources and tend to be useful only for synthesizing realistic reflections (known as ``reflection mapping''~\cite{blinn1976texture} in computer graphics). 

% While most of the literature has focused on more accurate HDR lighting estimation results~\cite{song2019neural,srinivasan2020lighthouse,li2020inverse,wang2021learning}, new approaches have started exploring another direction: editability. 2 recent techniques propose to do so: StyleLight~\cite{wang2022stylelight} (StyleGAN inversion) and Weber~\etal~\cite{weber2022editable} (single dominant parametric light source). Since HDR constraint (need accurate lighting), works only on indoors.

In this paper, we present a single, coherent framework that unifies these three main trends. Our approach, dubbed \thename, predicts a rich light representation in the form of a highly detailed \ang{360} environment map; is domain-generic as it works on both indoor and outdoor scenes seamlessly; and is editable, as it estimates individual HDR light sources from the image which can synthesize both realistic shading and reflections (see \cref{fig:teaser}). \thename is, to the best of our knowledge, the first editable HDR lighting estimation technique that works on both indoor and outdoor scenes seamlessly. Our work bridges the gap between HDR parametric lighting estimation and high-resolution field of view extrapolation by introducing a novel editable lighting co-modulation technique, which combines the flexibility and intuitiveness of parametric lighting models with the generative power of GANs. Extensive experiments demonstrate that \thename either compares favorably or outperforms indoor- and outdoor-specific approaches, both qualitatively and quantitatively.

% \cite{karimi2022guided}

%% file: 2_related_work.tex
\section{Related work}
\label{sec:rel_work}

\paragraph{Field of view extrapolation} Several image-based techniques~\cite{efros1999texture,efros2001image,barnes2009patchmatch,guillemot2013image} were proposed to extend images beyond their original frame by re-using their content. Since the inception of generative imaging~\cite{radford2015unsupervised,isola2017image,tewari2020state}, learning-based methods have delivered increasingly promising results for image inpainting and extrapolation tasks. Of note, CoModGAN~\cite{zhao2021comodgan}, proposes to convert unconditional generators such as StyleGAN~\cite{karras2019style} to conditional models by co-modulating both conditional inputs and the stochastic style representation throughout the generator. Karimi~\etal~\cite{karimi2022guided} extended this architecture to \ang{360} panorama extrapolation from regular field of view images. Similarly, Akimoto~\etal~\cite{akimoto2019360} suggest performing panorama extrapolation using a two-stage GAN and a transformer-based architecture~\cite{akimoto2022diverse}. Kulkarni~\etal \cite{kulkarni2022360fusionnerf} propose to use implicit radiance fields (NeRF \cite{mildenhall2020nerf}) coupled with a cross-domain embedding \cite{radford2021learning} to fill the occluded regions of the scene semantically. 

\paragraph{Dynamic range extrapolation} Cameras capture a limited dynamic range, resulting in saturated pixels when the radiance is outside this range. Over the years, methods have been developed \cite{eilertsen2017hdr,endo2017deep,marnerides2018expandnet,liu2020single,yu2021luminance} to recover the original values from those saturated pixels. Of note, Zhang and Lalonde~\cite{zhang-iccv-17} propose a method to extend the dynamic range for outdoor \ang{360} panoramas, recovering the correct sun intensity according to the weather conditions. 

\begin{figure*}[th]
\centering
\footnotesize
\includegraphics[width=\linewidth]{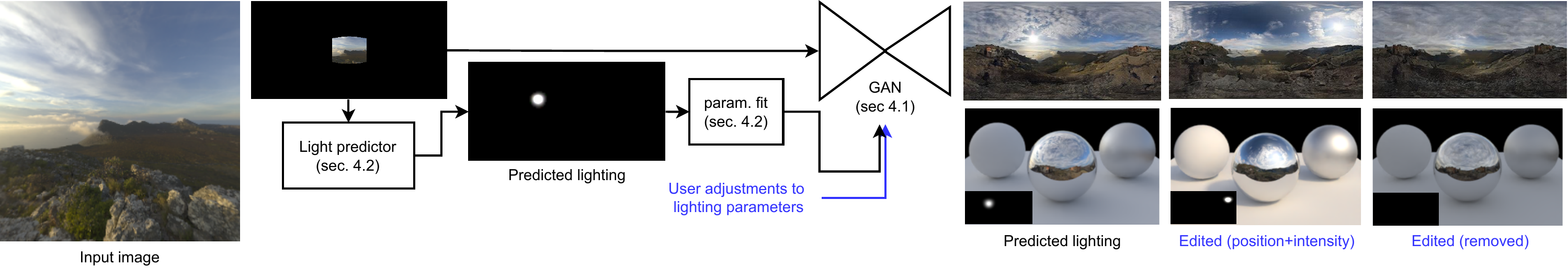}
\caption{Overview of our proposed approach. Our method accepts as input a single image (left), and produces a parametric lighting representation. The image and predicted lighting are both fed to a GAN which generates a highly detailed, high dynamic range environment map (right, top), which can be used to render virtual scenes realistically (right, bottom). The lighting representation can be intuitively edited by a user to produce controllable relighting results, such as moving the sun around (right, middle), or removing it entirely (far right). The parametric lighting representation is shown as insets on the renders. }
\label{fig:overview}
\end{figure*}

\paragraph{Single image lighting estimation and modeling} Illumination estimation was first approached for outdoor environments using explicit cues like detected shadows on the ground and shading on vertical walls \cite{lalonde2010sun,lalonde2014lighting}. Later, deep learning methods were developed to replace these explicit cues, leveraging either parametric sky models \cite{hold2017deep,zhang2019all, yu2021hierarchical} or non-parametric representations \cite{legendre2019deeplight,holdgeoffroy2019deep,zhu2021spatially,tang2022estimating}. 

% Indoor lighting conditions are more challenging than outdoor conditions due to additional uncertainties such as the number of lights, and their color and intensity, contrarily to outdoor environments where lighting is mainly explained by the well-studied sun and its interaction with the sky. 
Indoor lighting estimation has also received much attention in the literature, initially proposing to model the light as HDR panoramas or environment maps~\cite{gardner2017learning,song2019neural,somanath2021hdr,wang2022stylelight}. This representation has the advantage of providing high-quality textures for reflections. Spherical harmonics~\cite{ramamoorthi2001efficient,mandl2017learning,garon2019fast,zhao2020pointar} and spherical gaussians~\cite{gardner2019deep,li2020inverse,zhan2021emlight,zhan2021gmlight} are common alternatives. Lighting estimation has also been studied exhaustively as part of inverse rendering methods \cite{barron2014shape,shu2017neural,sengupta2019neural,li2020inverse,wang2021learning,wang2022neural,li2022physically,zhu2022irisformer}. The lighting representation of those methods is typically dense (either spatial or volumetric). 

\paragraph{Editable lighting estimation} All these aforementioned lighting representations, while more and more accurate, are typically intricate for users to edit. Notable exceptions are parametric models~\cite{hold2017deep,zhang2019all,gardner2019deep} that are intuitive but lack realism for reflections. Recently, Weber~\etal~\cite{weber2022editable} proposed to condition the texture generation with a single parametric light but are limited to indoor environments. Similarly, \cite{wang2022stylelight} can generate editable results, but rely on an expensive GAN inversion technique. In contrast, we propose an editable lighting estimation technique which generates high quality reflections, can handle more than one dominant light source, and requires a single forward pass in our network. 

% In our work, we leverage a parametric lighting representation named spherical gaussians, allowing easy user editing. Gardner~\etal \cite{gardner2019deep} were the first to employ deep learning to estimate indoor illumination as a fixed number of spherical gaussians from a regular image. 
% Later, methods were developed to extend this idea to a grid of gaussians on the whole sphere \cite{zhan2021emlight} or apply them to estimated 3D geometry \cite{zhan2021gmlight}. Inspired by these works, our method takes advantage of parametric spherical gaussians while keeping the high-resolution details provided by non-parametric panoramas. 

% Several other methods perform relighting 
% \cite{griffiths2022outcast}
% A new paradigm was recently proposed to relight in images using an implicit representation of the scene using GANs \cite{bhattad2022enriching} or diffusion models \cite{zhang2023adding}. While this new paradigm currently lacks fine-grained control over lighting, it provides promising results. 

%% file: 3_method.tex
\section{Background}
\subsection{Image formation}

As in \cite{somanath2021hdr}, we frame HDR lighting estimation as out-painting in a latitude-longitude (or equirectangular) panoramic representation. We warp the input image $\mathbf{I}$ to a \ang{360} panorama $\mathbf{X} \in \mathbb{R}^{H \times W \times 3}$, where $H$ and $W$ are the panorama height and width respectively ($W = 2H$), according to a simple pinhole camera model (with common assumptions: the principal point is the image center, negligible skew, unit pixel aspect ratio~\cite{Hartley2004}). We also assume knowledge of the camera parameters (field of view and camera elevation and roll) as in \cite{hara2021spherical,akimoto2019360,somanath2021hdr,karimi2022guided}. 
 
\subsection{Style co-modulation}

We base our approach on the style co-modulation framework of Zhao~\etal~\cite{zhao2021comodgan}, adapted to the case of \ang{360} fov extrapolation by Karimi~\etal~\cite{karimi2022guided}, which we briefly summarize here for completeness and illustrate in \cref{fig:method}a. The input, partially observed panorama $\mathbf{X}$ is given as input to an image encoder $\mathcal{E}_i$, whose output is combined to that of the mapper $\mathcal{M}$ via an affine transform $A$:
\begin{equation}
    \mathbf{w}^\prime = A\left(\mathcal{E}_i(\mathbf{X}), \mathcal{M}(\mathbf{z})\right) \,,
    \label{eq:co-modulation}
\end{equation}
where $\mathbf{z} \sim \mathcal{N}(0, \mathbf{I})$ is a random noise vector, and  $\mathbf{w}^\prime$ is the style vector modulating the generator $\mathcal{G}$. The output of $\mathcal{E}_i(\mathbf{X})$ is also provided as the input tensor to $\mathcal{G}$. Finally, the known portion of the input panorama $\mathbf{X}$ is composited over the output of the synthesis network $\hat{\mathbf{Y}}^\prime$ to obtain the final result $\hat{\mathbf{Y}}$ (we use the hat ($\,\hat{}\,$) notation to denote an output of our method). This ensures both the observed image and the user-edited lights are preserved by the method. 

\section{Editable lighting co-modulation}

\subsection{Method overview}

\begin{figure}[t]
    \centering
    \footnotesize
    \renewcommand{\tabcolsep}{3pt}
    \begin{tabular}{cc}
    \includegraphics[height=6.5cm,trim=0.4cm 0 1.1cm 0,clip]{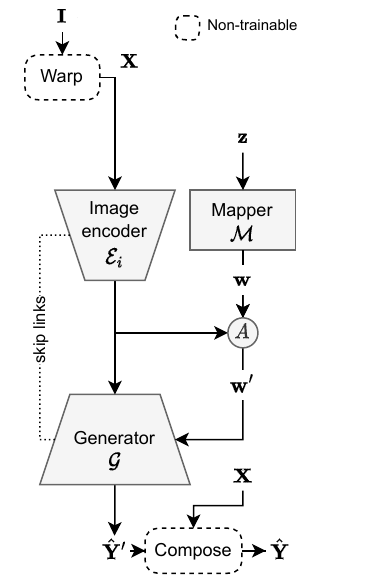} & 
    \includegraphics[height=6.5cm,trim=0.3cm 0 0.8cm 0,clip]{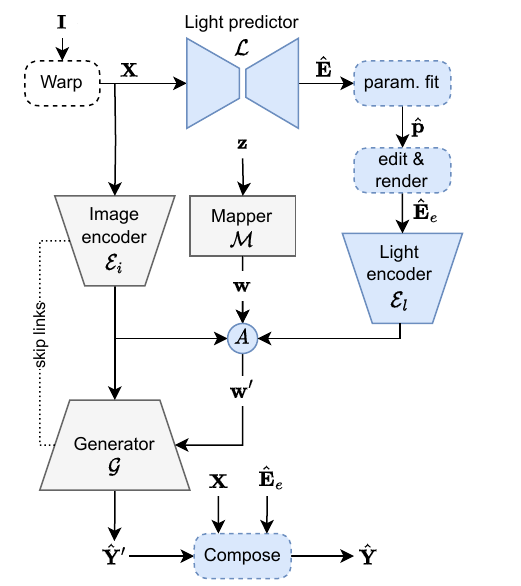} \\ 
    (a) Co-modulation~\cite{zhao2021comodgan,karimi2022guided} & (b) Editable lighting co-modulation \\
    \end{tabular}
    \caption[]{Overview of our proposed editable lighting co-modulation method. The input image $\mathbf{I}$ is first warped according to its (known) camera parameters to a \ang{360} panorama $\mathbf{X}$ via a (non-trainable) warp. (a)~As proposed in \cite{zhao2021comodgan,karimi2022guided}, an image encoder $\mathcal{E}_i$ co-modulates (with the mapper $\mathcal{M}$) the generator network $\mathcal{G}$ according to $\mathbf{X}$. Here, $A$ denotes a (learnable) affine transformation. (b)~Our editable light co-modulation method first estimates an HDR light map $\hat{\mathbf{E}}$ from the warped image $\mathbf{X}$ through a light predictor $\mathcal{L}$. The light map is converted to a parametric model $\hat{\mathbf{p}}$ and can optionally be edited by a user to obtain $\hat{\mathbf{p}}_e$ before being re-rendered to an image $\hat{\mathbf{E}}_e$ and injected in the style co-modulation mechanism via a light encoder $\mathcal{E}_l$. Our proposed novel components are highlighted in blue.}
    \label{fig:method}
\end{figure}

\Cref{fig:overview} presents a broad overview of our method for estimating a highly detailed, editable HDR lighting environment map from a single image. At the heart of our method is a novel editable lighting co-modulation mechanism, which is illustrated in greater detail in \cref{fig:method}b. A light prediction network $\mathcal{L}$ produces an HDR light environment map $\hat{\mathbf{E}} \in \mathbb{R}^{H \times W \times 3}$ from the input partially-observed panorama $\mathbf{X}$. The light map is then converted to a parametric lighting form $\hat{\mathbf{p}}$, which can, optionally, be intuitively edited by a user to obtain $\hat{\mathbf{p}}_e$. It is then rendered back to a panorama $\hat{\mathbf{E}}_e \in \mathbb{R}^{H \times W \times 3}$ before being fed to a light encoder $\mathcal{E}_l$, whose output is concatenated to the other vectors produced by the image encoder $\mathcal{E}_i$ and mapper $\mathcal{M}$ before being given to the affine transform $A$. This modulates the style injection mechanism of the generator with information from the input image, the random style, and the lighting information, hence the entire style co-modulation process becomes
\begin{equation}
    \mathbf{w}^\prime = A\left(\mathcal{E}_i(\mathbf{X}), \mathcal{M}(\mathbf{z}), \mathcal{E}_l(\hat{\mathbf{E}}_e) \right) \,.
    \label{eq:light-co-modulation}
\end{equation}
Finally, the (edited) light map $\hat{\mathbf{E}}_e$ is also composited with $\hat{\mathbf{Y}}^\prime$ to produce the final result $\hat{\mathbf{Y}}$. 

\subsection{Lighting representation prediction}
\label{sec:lighting-representation}

\paragraph{Parametric lighting representation $\mathbf{p}$} Similar to \cite{gardner2019deep,li2020inverse}, we model the dominant light sources in a scene as isotropic spherical gaussians. Given a set of $K$ spherical gaussians, light intensity $L(\omega)$ along (unit) direction vector $\omega \in \mathbb{S}^2$ is given by
\begin{equation}
L(\omega; \{\mathbf{c}_k, \xi_k, \sigma_k\}_{k=1}^K) = \sum_{k=1}^K \mathbf{c}_k G(\omega; \xi_k, \sigma_k) \,, 
\label{eq:sg}
\end{equation}
where $G(\omega; \xi, \sigma) = \exp\left(-\frac{1}{\sigma{}^2} (1-\omega \cdot \xi) \right)$. Here, $K$ denotes the number of individual light sources, $\mathbf{c}_k$ the RGB intensity of each light source. $\xi_k \in \mathcal{S}^2$ and $\sigma_k \in \mathbb{R}^1$ control the direction and bandwidth of each light source, respectively. Each light source is represented by three parameters $\mathbf{p} = \{\mathbf{c}_k, \xi_k, \sigma_k\}$. This compact, parametric form of spherical gaussians makes them suitable for editing: users can easily understand and modify their parameters. After editing, the spherical gaussians are rendered to an image format $\hat{\mathbf{E}}_e$ using \cref{eq:sg} before being given to the light encoder $\mathcal{E}_l$. 

\paragraph{Light predictor $\mathcal{L}$} Directly predicting multiple parametric lights requires a complex 2-stage training procedure~\cite{gardner2019deep}, which can be simplified if only a single dominant light is recovered~\cite{weber2022editable}. Here, we instead train a network $\mathcal{L}$ to predict the light sources in an image format (akin to the first stage of \cite{gardner2019deep} and \cite{somanath2021hdr}).

\paragraph{Spherical gaussian fitting} To obtain the parameters $\mathbf{p}$ from both the predicted light map $\hat{\mathbf{E}}$ and real panoramas $\mathbf{E}$, we employ the following procedure. We first threshold the HDR values on which we compute their connected components. We initialize the gaussians position $\xi_k$ and intensity at the center of mass and the maximum intensity of each connected component, respectively. We initialize all gaussian bandwidths with a fixed $\sigma = 0.45$. 
%Writing the spherical Gaussian parameters $\tau = \{c_k, \xi_k, \sigma_k\}_{k=1}^K$ for brevity, we obtain the  the minimize the L2 distance between the spherical Gaussian render $f_\mathrm{SG}(\omega; \tau )$ 
We obtain the light parameters $\mathbf{p}$ by optimizing the L2 reconstruction error over every pixel of the panorama $\Omega$ as
\begin{equation}
\hat{\mathbf{p}} = \argmin_{\mathbf{p}} \sum_{\omega \in \Omega} \lambda_1 \left\lVert L(\omega; \mathbf{p}) - \mathbf{E}(\omega) \right\rVert_2^2 + \ell_\mathrm{reg}( \mathbf{p} ) \,,
\end{equation}
where $\lambda_1$ acts as a loss scaling factor and $\ell_\mathrm{reg}( \mathbf{p} )$ is a regularizing term stabilizing the optimization over light vector length, intensity, bandwidth, and color (see supplementary for more details). We use non-maximal suppression to fuse overlapping lights during the optimization. 

\subsection{Data}

To train our light predictor and generator, we employ a dataset consisting of \ang{360} real captured panoramas purchased from 360cities\footnote{\url{https://360cities.net}}. We split our dataset into $239\,064$, $1000$, and $1000$ different panoramas for train, validation, and test purposes, respectively. We extend the dynamic range of the panoramas to HDR using the method of Zhang et al.~\cite{zhang-iccv-17} trained on the Laval indoor, outdoor, and sky HDR databases \cite{skydb,gardner2017learning,holdgeoffroy2019deep}.

\subsection{Implementation details}

We implement our lighting co-modulation, light predictor, and spherical gaussian fitting algorithm using PyTorch \cite{paszke2019pytorch}. 
We train our image encoder $\mathcal{E}_i$, mapper $\mathcal{M}$, light encoder $\mathcal{E}_l$, and panorama generator $\mathcal{G}$ simultaneously using Adam with a learning rate of 0.002 for four days on eight A100 GPUs.
We implement our light predictor with a UNet with fixup initialization~\cite{griffiths2022outcast,zhang2019fixup} for 12 hours on four A100 GPUs. The light predictor network contains five downsampling layers and one bottleneck layer. We use the cosine blurring filter of \cite{gardner2017learning} and compute the losses in log domain to further stabilize the training of our light predictor.
Our gaussian fitting algorithm optimizes the light parameters $\mathbf{p}$ using stochastic gradient descent without momentum and a learning rate of $5\times{}10^{-4}$. We leverage a learning rate reduction strategy that reduces it by two for every 20 epochs the loss did not improve. For its initialization, we first blur the panorama with a gaussian filter $\sigma=3$, then threshold it to its 98.5\textsuperscript{th} percentile and use $\lambda_1 = \nicefrac{1}{50}$. 

%% file: 4_experiment.tex
\section{Experiments}
\label{sec:experiments}

We now evaluate our \thename method against the recent state of the art. Because most previous methods are designed either for indoor or outdoor lighting, we separate the experiments accordingly. One notable exception is the work of Legendre~\etal~\cite{legendre2019deeplight} which works across both domains---however, since neither code nor data are available we unfortunately could not include it in the evaluation. 

\subsection{Evaluation on indoor images}
\label{sec:evaluation-indoor}

\paragraph{Quantitative comparison} We first demonstrate that our model performs either on par or better than the state of the art on quantitative metrics evaluated on indoor images. For the evaluation protocol, we follow \cite{weber2022editable} and rely on the test set provided by \cite{gardner2017learning}. For each of the 224 panoramas in the test split, we extract 10 LDR images using the same sampling distribution as in \cite{gardner2017learning}, for a total of 2,240 images for evaluation. Each image is given as input to a technique, and the resulting lighting representation is used to render a virtual scene composed of 9 diffuse spheres on a ground plane, seen from above (see \cite{weber2022editable}). The following metrics are then computed on the resulting renders: RMSE and its scale-invariant version (siRMSE)~\cite{barron2014shape}, RGB ang.~\cite{legendre2019deeplight}, and PSNR. In addition, we also report the FID~\cite{heusel2017gans} computed directly on the lighting representation expressed in equirectangular format. To avoid overly favoring techniques trained and evaluated on the Laval Indoor HDR Dataset~\cite{gardner2017learning}, we extend the evaluation from \cite{weber2022editable} and compute the FID against a test set of 1,093 unique indoor panoramas including the Laval Indoor HDR test set~\cite{gardner2017learning} (305), indoor panoramas extracted from \cite{cheng2018shlight} (192), and 360cities test set \cite{karimi2022guided} (596).

We compare \thename to the following methods. Two versions of \cite{gardner2019deep} are compared: the original (3) where three light sources are estimated, and a version (1) trained to predict a single parametric light. We also compare to Lighthouse~\cite{srinivasan2020lighthouse}, which expects a stereo pair as input, but we generate a second image with a small baseline using \cite{wiles2020synsin} (visual inspection confirmed this yields results comparable to the published work). For \cite{garon2019fast}, we select the coordinates of the image center for the object position. For \cite{somanath2021hdr}, we implemented their proposed ``Cluster ID loss'', and tonemapping but used pix2pixHD~\cite{wang2018high} as backbone. We compare against EMLight~\cite{zhan2021emlight} and StyleLight~\cite{wang2022stylelight} with the provided code. Finally, we also include \cite{weber2022editable}, and the recent work of \cite{karimi2022guided} as a state-of-the-art (LDR) field of view extrapolation method.

\Cref{tab:quantitative-indoor} shows the quantitative comparison results on indoor scenes. Our method achieves a strong balance between rendering scores (left) and FID (right). Notably, its FID is only slightly above (78.90 vs 65.98) that of ImmerseGAN~\cite{karimi2022guided}, which shows that adding the controllable spherical gaussian lighting on the generated panoramas does not significantly affect their realism. While our results are on par with StyleLight~\cite{wang2022stylelight}, our feed-forward method does not require any time-consuming GAN inversion process, which takes around 0.07 seconds on a GeForce RTX 2070 GPU (versus 58 seconds for StyleLight~\cite{wang2022stylelight}). 
Furthermore, the next best methods have much higher FID: Weber'22~\cite{weber2022editable} with 130.13, and EMLight ~\cite{zhan2021emlight} with 135.97.
\input{tables/quantitative_indoor_table}

\paragraph{Qualitative evaluation}
We present qualitative results in \cref{fig:qual-indoor}, where a virtual scene rendered with the corresponding lighting predictions for each technique are shown. To best illustrate the impact of HDR lighting (shading, shadows) and reflections, we render a scene composed of 3 spheres with varying reflectance properties (diffuse, mirror, glossy) on a flat diffuse ground plane. In addition, a tonemapped equirectangular view of the estimated light representation is provided under each render. While ImmerseGAN~\cite{karimi2022guided} yields highly realistic reflections, all renders obtained with this method are devoid of contrast, strong shading, and shadows since its output is LDR. The quality of the shading and shadows obtained with \thename is qualitatively similar to those of StyleLight~\cite{wang2022stylelight} and Weber'22~\cite{weber2022editable}, despite these techniques being trained solely for indoor images, on the Laval Indoor HDR dataset. 

\input{tables/qualitative_indoor}

\subsection{Evaluation on outdoor images}
\label{sec:evaluation-outdoor}
We now evaluate on outdoor environments. Note that our method has not changed: the exact same \thename model works interchangeably on indoor and outdoor images. 

\paragraph{Quantitative evaluation} Here, we rely on the outdoor panoramas from the \cite{cheng2018shlight} dataset, which contains 839 unique outdoor panoramas. For each panorama in this set, we extract three LDR images with the azimuth of $h_\theta \in \{0, 120, 240\}^\circ$, for a total of 2,517 images for evaluation. As in \cref{sec:evaluation-indoor}, each image is given as input to a technique, and the resulting lighting representation is used to render the same virtual scene. The same metrics are reported as well, except that the FID is computed on the outdoor test set described above. This time, we compare against the work of Zhang~\etal~\cite{zhang2019all}, who proposed a method for predicting the parameters of an outdoor sun+sky lighting model~\cite{lalonde2014lighting}. We also include ImmerseGAN~\cite{karimi2022guided} as it was trained to extrapolate the field of view of outdoor images as well. \Cref{tab:quantitative-outdoor} reports the quantitative comparison results on outdoor scenes. Here, our method vastly outperforms the previous outdoor-specific technique of Zhang'19~\cite{zhang2019all} on all metrics. 
% In addition, it reports an FID of 61.49, similar to its indoor performance, demonstrating a strong consistency. 

\input{tables/quantitative_outdoor_table}

\paragraph{Qualitative evaluation} Corresponding qualitative results are presented in \cref{fig:qual-outdoor}, where a virtual scene rendered with the corresponding lighting predictions for each technique are shown. While the lighting representation employed by \cite{zhang2019all} is able to produce very strong shadows, the predicted lighting always results in almost-constant gray skies, which results in somewhat neutral renderings. ImmerseGAN~\cite{karimi2022guided}, on the other hand, creates much more lively renderings, which however lack shadows and shading. Our method bridges these two works and offers results which combine the best of both worlds: realistic reflections with more pronounced shading and cast shadows. 

\input{tables/qualitative_outdoor}

\subsection{Editing the estimated lighting}
Because our approach estimates a set of parametric light sources as spherical gaussians (c.f. \cref{sec:lighting-representation}), a user can intuitively edit their properties before generating the panorama with the generator. We demonstrate editing capability by showing combinations of input images and desired lighting in \cref{fig:qual-imgxsg}. In all cases, the desired spherical gaussians are realistically blended with the generated panoramas, sometimes even creating compelling interactions such as reflections on the ground or the sea. We demonstrate additional editing capabilities such as removing and/or adding light sources in \cref{fig:qual-editing}. Compared to StyleLight~\cite{wang2022stylelight}, our method requires a single forward pass in the network as opposed to a computationally expensive GAN inversion step. More importantly, our method works equally well for indoor and outdoor images, as shown in both \cref{fig:qual-editing,fig:qual-imgxsg}.

\input{tables/qualitative_edit}
\input{tables/qualitative_IMGxSG}

\subsection{Ablation studies}
\begin{figure}[t!]
    \centering
    \renewcommand{\tabcolsep}{1pt}
    \newcommand{\mywidth}{0.5\linewidth}
    \resizebox{\linewidth}{!}{% take the[] entire width, but still find a good per-image width otherwise text gets compressed
    \begin{tabular}{ccc}
    Lighting parameters & No lighting co-modulation & Ours \\
    \includegraphics[width=\mywidth]{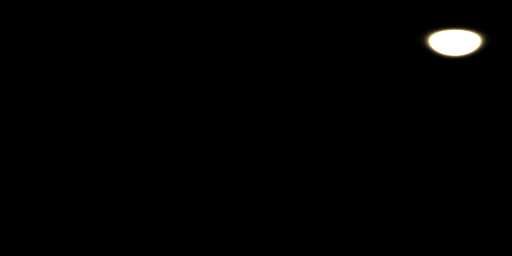}
    & \includegraphics[width=\mywidth]{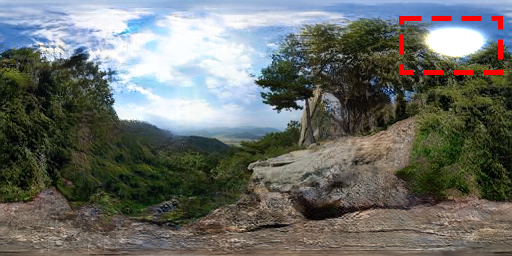}
    & \includegraphics[width=\mywidth]{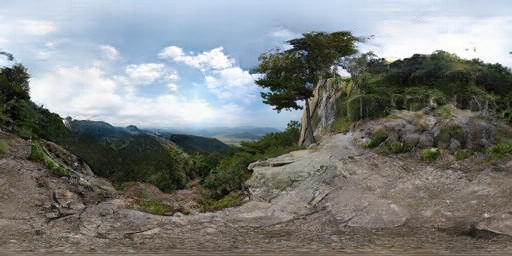}
    \\
    \includegraphics[width=\mywidth]{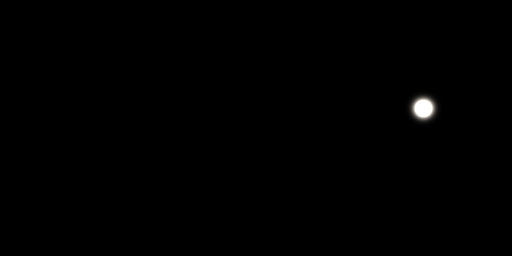}
    & \includegraphics[width=\mywidth]{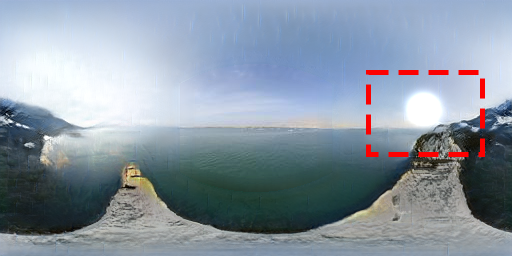}
    & \includegraphics[width=\mywidth]{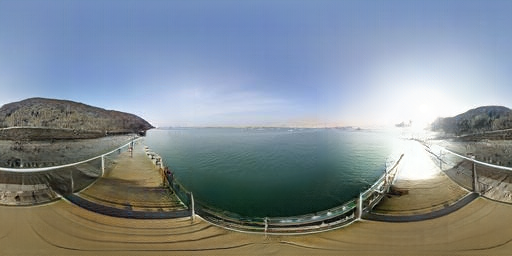}
    \\
    \end{tabular}}
    \caption[]{Ablation study on the effect of our proposed lighting co-modulation. Using the lighting co-modulation helps blending realistically the lighting parameters into the surrounding.}
    \label{fig:ablation}
\end{figure}

% Karimi et al.~\cite{karimi2022guided} performed an ablation study where they used Pix2Pix-HD~\cite{wang2018high} as a feed-forward GAN without co-modulation. They show this leads to mode-collapse-like visual artifacts. While Karimi et al.~\cite{karimi2022guided} achieve the state-of-the-art results for field of view extrapolation, using their method leads to a limited dynamic range. In contrast, our proposed light predictor estimates high dynamic range light sources. To further blend these light sources realistically into a \ang{360} panorama, we proposed lighting co-modulation. To 
%  evaluate the impact of our proposed lighting co-modulation, we perform an ablation study by removing it and adding estimated lighting parameters directly to the output of the generator. As shown in \cref{fig:ablation}, removing the lighting co-modulation prevents the generator from realistically blending the lighting parameters with its surroundings.
To evaluate the impact of our proposed lighting co-modulation, we perform an ablation study by removing the lighting encoder ($\mathcal{E}_l$ in \cref{fig:method}) and compositing the estimated lighting directly to the generator output. In \cref{tab:ablation}, we report the FID computed on the same test sets as in \cref{sec:evaluation-indoor,sec:evaluation-outdoor}. While removing the lighting co-modulation does not have a noticeable impact for indoor images, in the outdoor case it prevents the generator from realistically blending the lighting parameters with its surroundings, as illustrated qualitatively in \cref{fig:ablation}. Note that we do not perform an ablation on the style co-modulation process from the mapper $\mathcal{M}$ (\cref{fig:method}) since \cite{karimi2022guided} show using a feed-forward GAN without co-modulation~\cite{wang2018high} leads to mode-collapse-like visual artifacts.

\begin{table}[t]
\footnotesize
\centering
\caption{Quantitative comparison between our proposed method with and without lighting co-modulation process in terms of FID. }
\begin{tabular}{lcc}
\toprule
& \multicolumn{2}{c}{FID$_\downarrow$} \\
& outdoor & indoor \\
\midrule
Ours & 38.44 & 78.90 \\
% Ours & 44.69 & 46.72\\ 
No lighting co-modulation & 50.29 & 79.47 \\
\bottomrule
\end{tabular}
\label{tab:ablation}
\end{table}

%% file: tables/quantitative_indoor_table.tex
\begin{table}[t]
\footnotesize
\setlength{\tabcolsep}{0.1pt}
\centering
\caption{Quantitative comparative metrics for indoor images. The metrics si-RMSE, RMSE, RGB ang., and PSNR (left) are computed on (tonemapped) renderings of a diffuse scene, and FID (right) on the estimated environment maps directly. Each row is color-coded as \colorbox{best}{best} and \colorbox{second}{second best}. We also \colorbox{best}{highlight} the methods which produce lighting representations that can be interpreted and edited by a user (``Edit.''). }

\label{tab:quantitative-indoor}
\begin{tabular}{lcccc}
\toprule
& si-RMSE$_\downarrow$
& RMSE$_\downarrow$
& RGB ang.$_\downarrow$
& PSNR$_\uparrow$ \\
\midrule
\thename (ours)     & 
\cellcolor{second}{0.091} & 0.238 & 6.36$^\circ$ & 10.03 \\ 
StyleLight~\cite{wang2022stylelight}      
& 0.123 & 0.316 & 7.09$^\circ$ & \cellcolor{second}{12.35} \\ 
Weber'22~\cite{weber2022editable}       
& \cellcolor{best}{0.079} & \cellcolor{best}{0.196} & \cellcolor{second}{4.08$^\circ$} & \cellcolor{best}{12.95} \\ 
Gardner'19 (1)~\cite{gardner2019deep} 
& 0.099 & 0.229 & 4.42$^\circ$ & 12.21  \\ 
Gardner'19 (3)~\cite{gardner2019deep} 
& 0.105 & 0.507 & 4.59$^\circ$ & 10.90  \\ 
Gardner'17~\cite{gardner2017learning} 
& 0.123 & 0.628 & 8.29$^\circ$ & 10.22 \\ 
Garon'19~\cite{garon2019fast}
& 0.096 & 0.255 & 8.06$^\circ$ & 9.73  \\ 
Lighthouse~\cite{srinivasan2020lighthouse} 
& 0.121 & 0.254 & 4.56$^\circ$ & 9.81 \\
EMLight~\cite{zhan2021emlight} 
& 0.099 & 0.232 & \cellcolor{best}{3.99$^\circ$} & 10.34 \\
EnvmapNet$^\dag$ ~\cite{somanath2021hdr} 
& {0.097} & {0.286} & {7.67$^\circ$} & {11.74} \\
ImmerseGAN~\cite{karimi2022guided}      
& 0.094 & \cellcolor{second}{0.226} & 8.61$^\circ$ & 10.72 \\ 
% \bottomrule
% dumb test      
% & 0.089 & 0.214 & 8.03$^\circ$ & 10.91 \\ 
\bottomrule
\end{tabular}
\begin{tabular}{c}
\toprule
FID$_\downarrow$ \\ 
\midrule
\cellcolor{second}{78.90} \\
\cellcolor{second}{78.55} \\
130.13 \\ 
410.12 \\ 
386.43 \\ 
253.40 \\ 
324.51 \\ 
174.52 \\ 
135.97 \\
221.85 \\
\cellcolor{best}{65.98} \\
\bottomrule
% INF \\
% \bottomrule
\end{tabular}
\begin{tabular}{c}
\toprule
Edit. \\ 
\midrule
\cellcolor{best}{yes} \\
\cellcolor{best}{yes} \\
\cellcolor{best}{yes} \\
\cellcolor{best}{yes} \\
\cellcolor{best}{yes} \\
no \\
no \\
no \\
no \\
no \\
no \\
% \bottomrule
% NA \\
\bottomrule
\end{tabular}
\\
{\scriptsize $^\dag$ Only their proposed ClusterID loss and tonemapping.}
\end{table}

%% file: tables/qualitative_indoor.tex
\begin{figure*} [t]
    \centering
    \renewcommand{\tabcolsep}{1pt}
    \newcommand{\mywidth}{0.15\linewidth}
    \resizebox{\linewidth}{!}{% take the[] entire width, but still find a good per-image width otherwise text gets compressed
    \begin{tabular}{ccccccccccc}
    Input image & GT & Gardner'17~\cite{gardner2017learning} & Gardn.'19 (3)*~\cite{gardner2019deep} & EMLight~\cite{zhan2021emlight} & Lighthouse~\cite{srinivasan2020lighthouse} & Garon'19~\cite{garon2019fast} & ImmerseGAN~\cite{karimi2022guided} & StyleLight*~\cite{wang2022stylelight} & Weber'22*~\cite{weber2022editable} & \thename (ours)* \\
    \includegraphics[width=\mywidth]{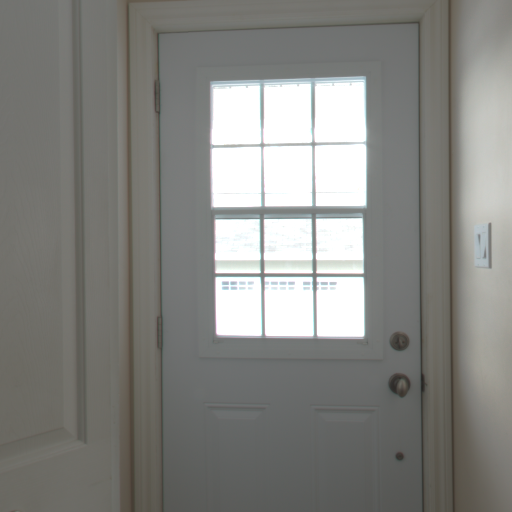} &
    \includegraphics[width=\mywidth]{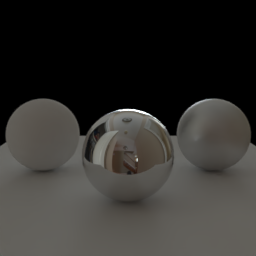} &
    \includegraphics[width=\mywidth]{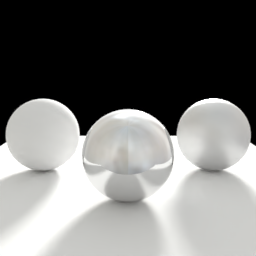} &
    \includegraphics[width=\mywidth]{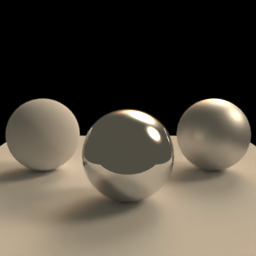} &
    \includegraphics[width=\mywidth]{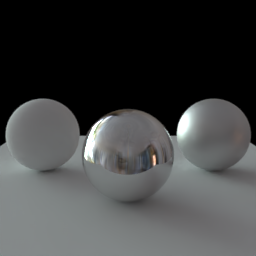} &
    \includegraphics[width=\mywidth]{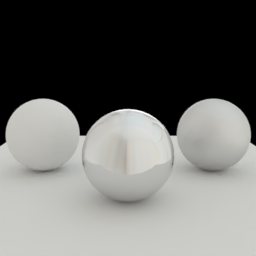} &
    \includegraphics[width=\mywidth]{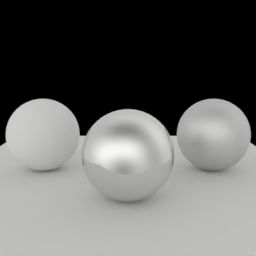} &
    \includegraphics[width=\mywidth]{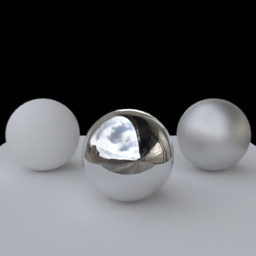} &
    \includegraphics[width=\mywidth]{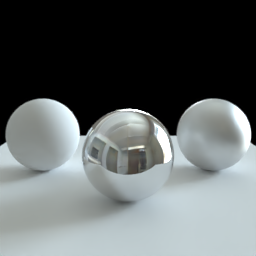} &
    \includegraphics[width=\mywidth]{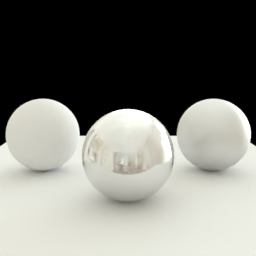} &
    \includegraphics[width=\mywidth]{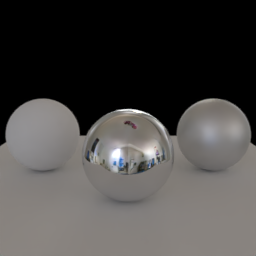}
    \\
    
    &
    \includegraphics[width=\mywidth]{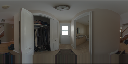} & 
     \includegraphics[width=\mywidth]{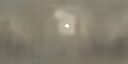} & 
     \includegraphics[width=\mywidth]{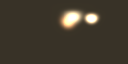} & 
     \includegraphics[width=\mywidth]{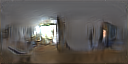} & 
     \includegraphics[width=\mywidth]{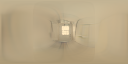} & 
     \includegraphics[width=\mywidth]{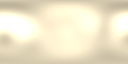} & 
     \includegraphics[width=\mywidth]{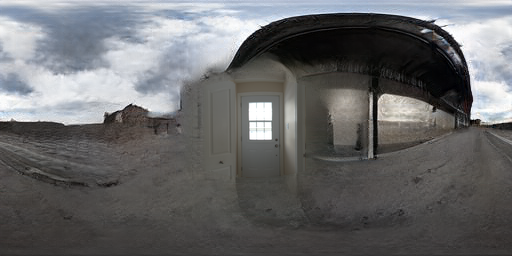} & 
     \includegraphics[width=\mywidth]{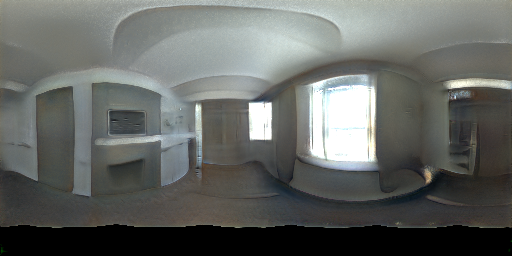} &
     \includegraphics[width=\mywidth]{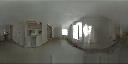} &
    \includegraphics[width=\mywidth]{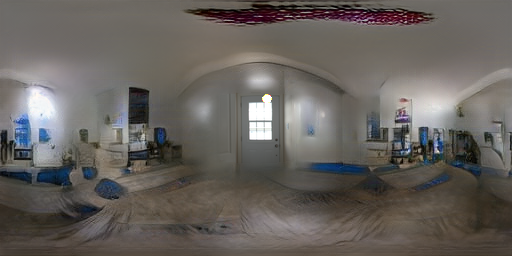}
    \\
    
    \includegraphics[width=\mywidth]{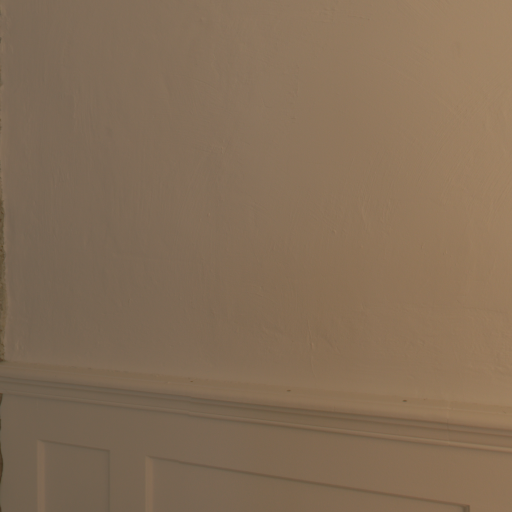} &
    \includegraphics[width=\mywidth]{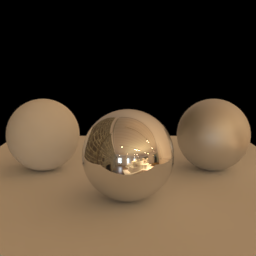} &
     \includegraphics[width=\mywidth]{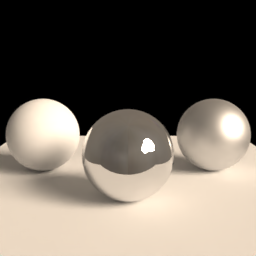} &
     \includegraphics[width=\mywidth]{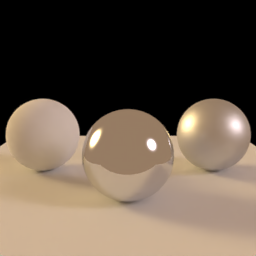} &
    \includegraphics[width=\mywidth]{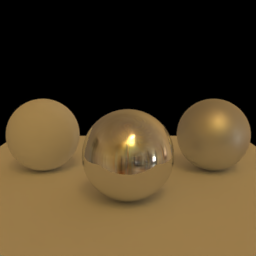} &
     \includegraphics[width=\mywidth]{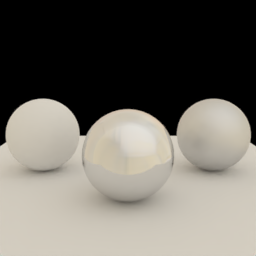} &
     \includegraphics[width=\mywidth]{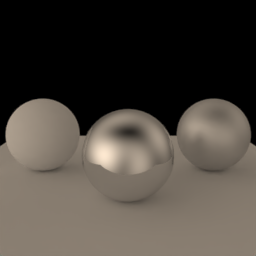} &
     \includegraphics[width=\mywidth]{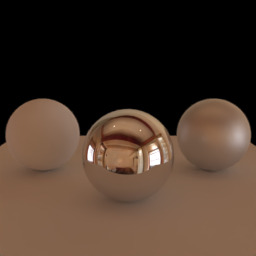} &
    \includegraphics[width=\mywidth]{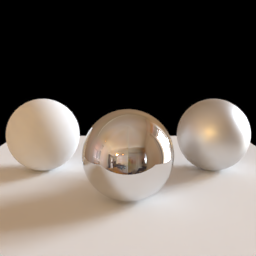} &
    \includegraphics[width=\mywidth]{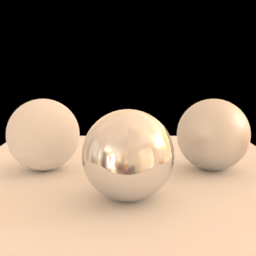} &
    \includegraphics[width=\mywidth]{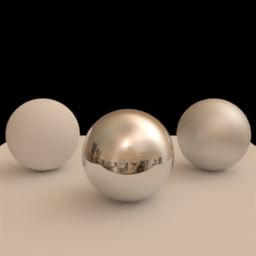}
    \\

     &
    \includegraphics[width=\mywidth]{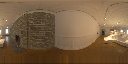} & 
     \includegraphics[width=\mywidth]{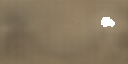} & 
     \includegraphics[width=\mywidth]{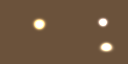} & 
      \includegraphics[width=\mywidth]{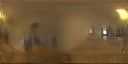} & 
      \includegraphics[width=\mywidth]{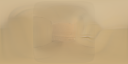} & 
      \includegraphics[width=\mywidth]{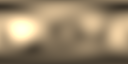} & 
      \includegraphics[width=\mywidth]{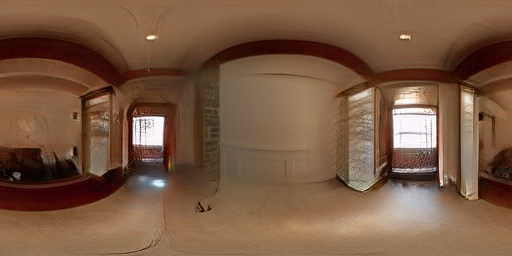} & 
      \includegraphics[width=\mywidth]{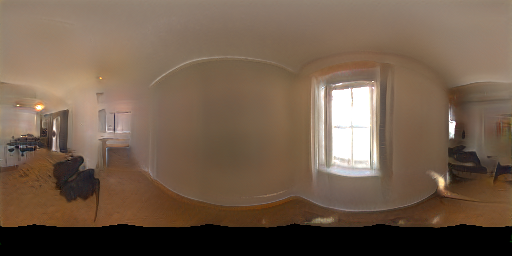} &
      \includegraphics[width=\mywidth]{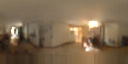} &
    \includegraphics[width=\mywidth]{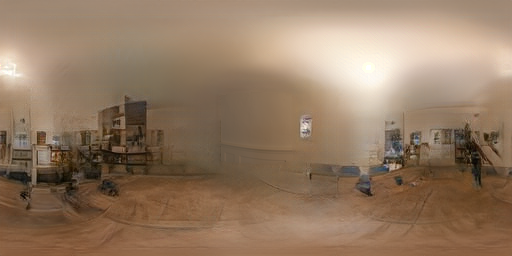}
    \\
    
    \includegraphics[width=\mywidth]{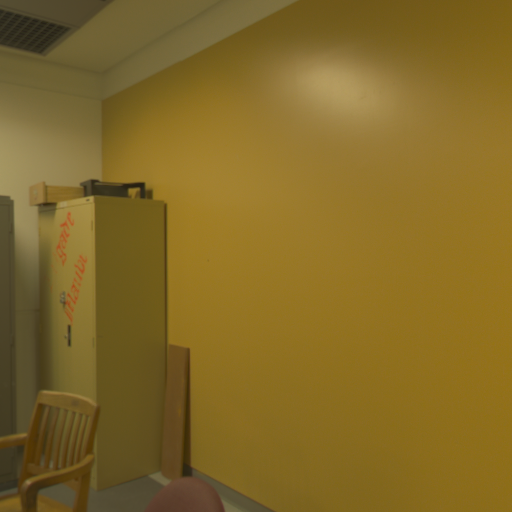} &
    \includegraphics[width=\mywidth]{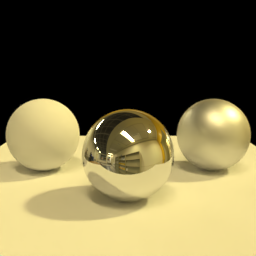} &
     \includegraphics[width=\mywidth]{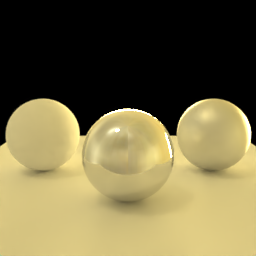} &
     \includegraphics[width=\mywidth]{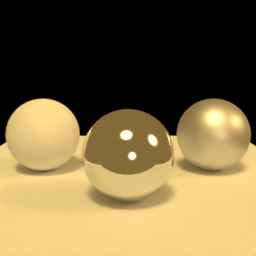} &
    \includegraphics[width=\mywidth]{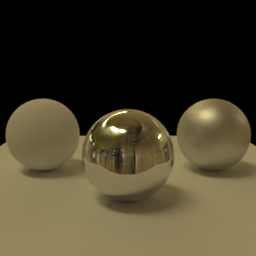} &
    \includegraphics[width=\mywidth]{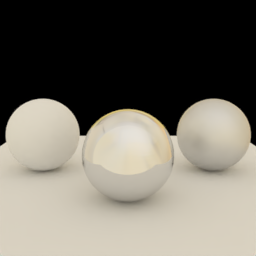} &
    \includegraphics[width=\mywidth]{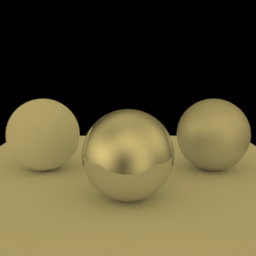} &
    \includegraphics[width=\mywidth]{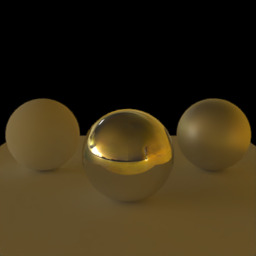} &
    \includegraphics[width=\mywidth]{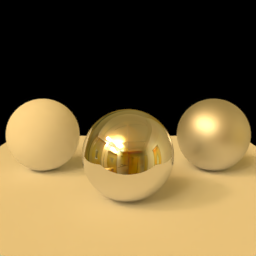} &
    \includegraphics[width=\mywidth]{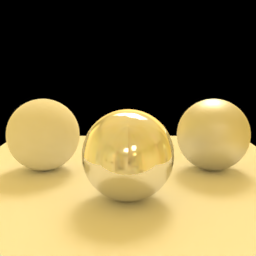} &
    \includegraphics[width=\mywidth]{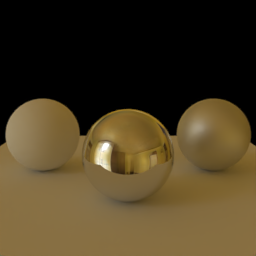}
    \\

    &
    \includegraphics[width=\mywidth]{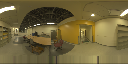} & 
     \includegraphics[width=\mywidth]{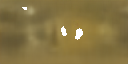} & 
     \includegraphics[width=\mywidth]{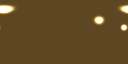} & 
     \includegraphics[width=\mywidth]{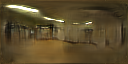} & 
     \includegraphics[width=\mywidth]{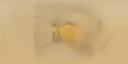} & 
     \includegraphics[width=\mywidth]{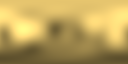} & 
     \includegraphics[width=\mywidth]{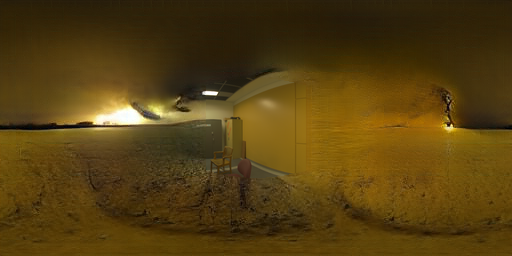} & 
     \includegraphics[width=\mywidth]{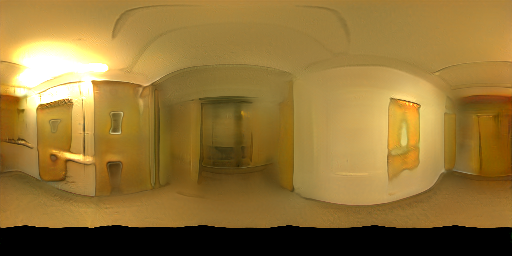} &
     \includegraphics[width=\mywidth]{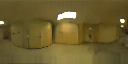} &
    \includegraphics[width=\mywidth]{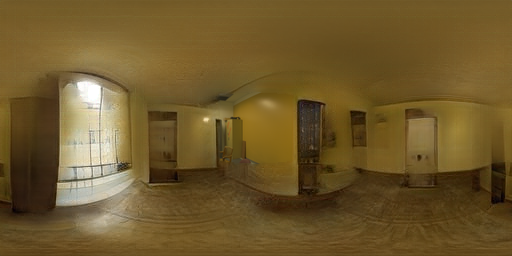}
    \\
    
     \includegraphics[width=\mywidth]{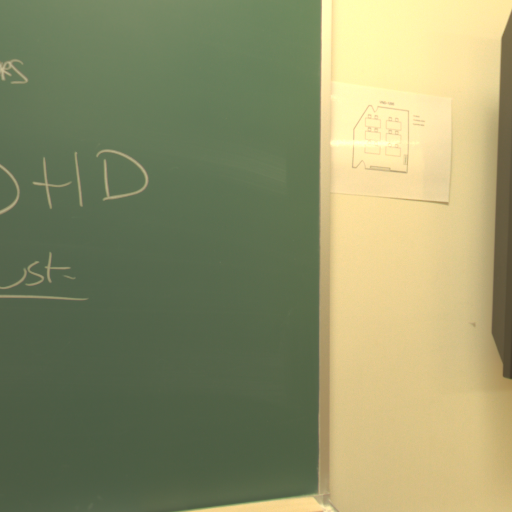} &
     \includegraphics[width=\mywidth]{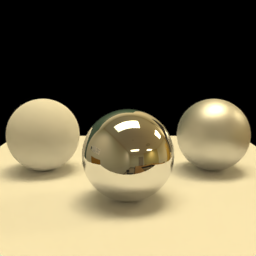} &
     \includegraphics[width=\mywidth]{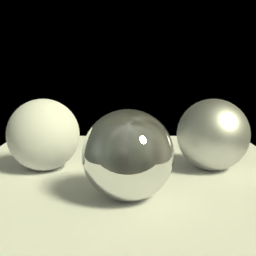} &
     \includegraphics[width=\mywidth]{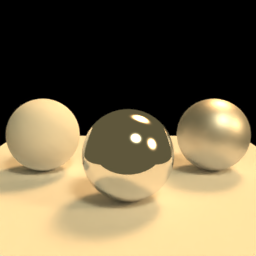} &
     \includegraphics[width=\mywidth]{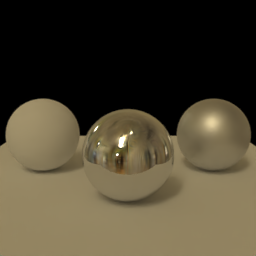} &
     \includegraphics[width=\mywidth]{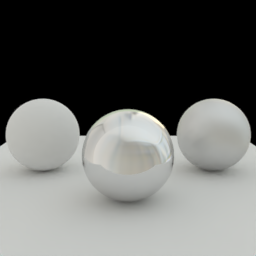} &
     \includegraphics[width=\mywidth]{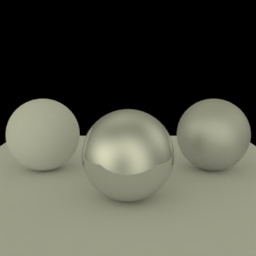} &
     \includegraphics[width=\mywidth]{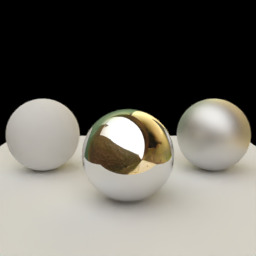} &
    \includegraphics[width=\mywidth]{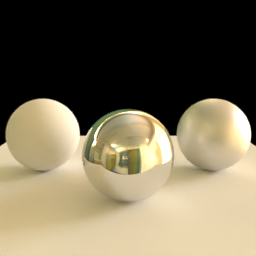} &
    \includegraphics[width=\mywidth]{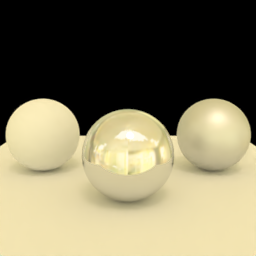} &
    \includegraphics[width=\mywidth]{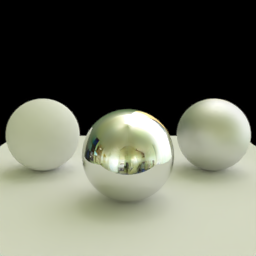}
    \\

    &
    \includegraphics[width=\mywidth]{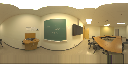} & 
    \includegraphics[width=\mywidth]{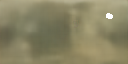} & 
    \includegraphics[width=\mywidth]{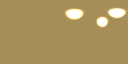} &  
    \includegraphics[width=\mywidth]{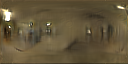} & 
    \includegraphics[width=\mywidth]{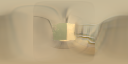} & 
    \includegraphics[width=\mywidth]{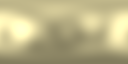} & 
     \includegraphics[width=\mywidth]{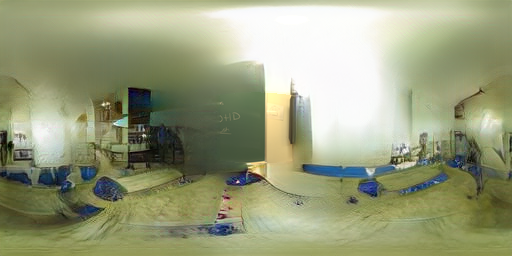} &
    \includegraphics[width=\mywidth]{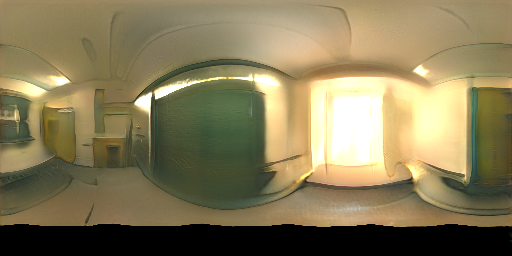} &
    \includegraphics[width=\mywidth]{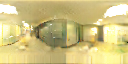} &
    \includegraphics[width=\mywidth]{figs/camera_ready/panos/9C4A0358-ffcdfa74fe_06.png}
    \\
    \end{tabular}}
    \vspace{0.25em}
    \caption[]{Qualitative rendering results for indoor scenes. We compare against the recent literature on single image lighting estimation, from left to right: Gardner~\etal~2017~\cite{gardner2017learning}, Gardner~\etal~2019 (with 3 parametric lights)~\cite{gardner2019deep}, EMLight~\cite{zhan2021emlight}, Lighthouse~\cite{srinivasan2020lighthouse}, Garon~\etal~2019~\cite{garon2019fast}, ImmerseGAN~\cite{karimi2022guided}, StyleLight~\cite{wang2022stylelight}, and Weber~\etal~2022~\cite{weber2022editable}. Our results are shown in the last column. All methods with editable lighting capabilities are indicated with *. From a given input crop (left), we show a render of a scene composed of three spheres (diffuse, mirror, glossy) on a diffuse ground plane (first row), and the corresponding estimated lighting (in equirectangular format) below.}
    \label{fig:qual-indoor}
\end{figure*}

%% file: tables/quantitative_outdoor_table.tex
\begin{table}[t]
\footnotesize
\setlength{\tabcolsep}{0.1pt}
\centering
\caption{Quantitative comparative metrics for outdoor images. The metrics si-RMSE, RMSE, RGB ang., and PSNR (left) are computed on (tonemapped) renderings of a diffuse scene, and FID (right) on the estimated environment maps directly. Each row is color-coded as \colorbox{best}{best} and \colorbox{second}{second best}. We also \colorbox{best}{highlight} the methods which produce lighting representations that can be interpreted and edited by a user (``Edit.''). }
\label{tab:quantitative-outdoor}
\begin{tabular}{lcccc}
\toprule
& si-RMSE$_\downarrow$
& RMSE$_\downarrow$
& RGB ang.$_\downarrow$
& PSNR$_\uparrow$ \\
\midrule
\thename (ours)      
& \cellcolor{best}{0.163} & \cellcolor{second}{0.469} & \cellcolor{best}{{8.53$^\circ$}} & \cellcolor{second}{10.03} \\ 
Zhang'19~\cite{zhang2019all}      
& 0.225 & 1.058 & {11.80$^\circ$} & 5.31 \\ 
ImmerseGAN~\cite{karimi2022guided}      
& \cellcolor{second}{0.174} & \cellcolor{best}{0.332} & \cellcolor{second}{{9.26$^\circ$}} & \cellcolor{best}{11.02} \\ 
\bottomrule
% dumb test      
% & 0.201 & 0.430 & 11.92 & 9.65 \\ 
% \bottomrule
\end{tabular}
\begin{tabular}{c}
\toprule
FID$_\downarrow$ \\ 
\midrule
\cellcolor{second}{38.44} \\
449.49\\
\cellcolor{best}{37.05}\\
\bottomrule
% INF\\
% \bottomrule
\end{tabular}
\begin{tabular}{c}
\toprule
Edit. \\ 
\midrule
\cellcolor{best}{yes} \\
\cellcolor{best}{yes} \\
no \\
% \bottomrule
% NA \\
\bottomrule
\end{tabular}
\end{table}

%% file: tables/qualitative_outdoor.tex
\begin{figure}[t]
    \centering
    \renewcommand{\tabcolsep}{1pt}
    \newcommand{\mywidth}{0.3\linewidth}
    \resizebox{\linewidth}{!}{% take the[] entire width, but still find a good per-image width otherwise text gets compressed
    \begin{tabular}{ccccc}
    Input & GT & Zhang'19*~\cite{zhang2019all} & ImmerseGAN~\cite{karimi2022guided} & \thename (ours)*   \\
    \includegraphics[width=\mywidth]{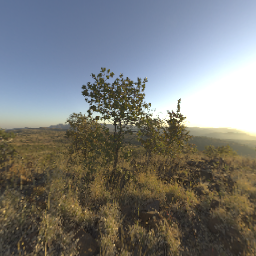}
    & \includegraphics[width=\mywidth]{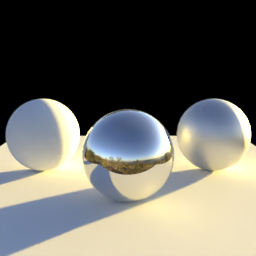}
    & \includegraphics[width=\mywidth]{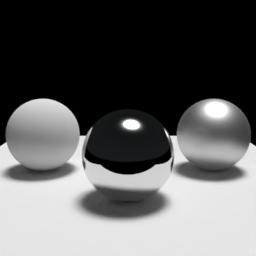}
    % & \includegraphics[width=\mywidth]{figs/qualitative_outdoor/render/immersegan/lenong_2_8k_00.png}
    & \includegraphics[width=\mywidth]{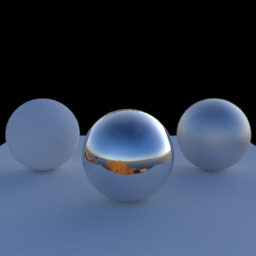}
    % & \includegraphics[width=\mywidth]{figs/qualitative_outdoor/render/everlight/lenong_2_8k_00.png}
    & \includegraphics[width=\mywidth]{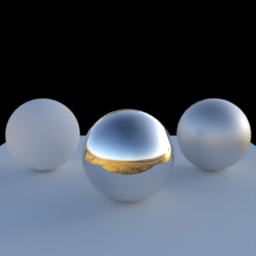}
    \\
    & \includegraphics[width=\mywidth]{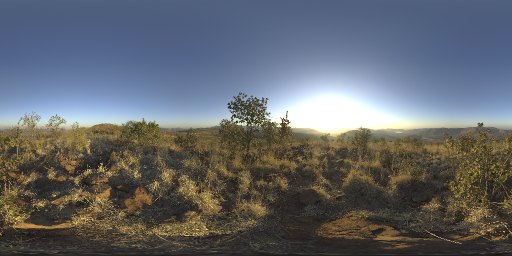}
    & \includegraphics[width=\mywidth]{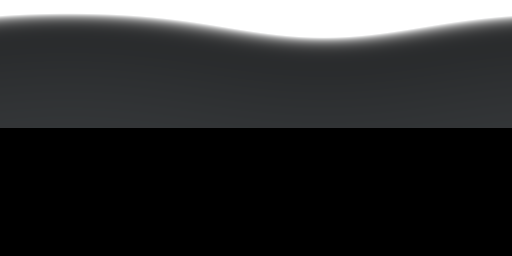}
    % & \includegraphics[width=\mywidth]{figs/qualitative_outdoor/pano/immersegan/lenong_2_8k_00.png}
    & \includegraphics[width=\mywidth]{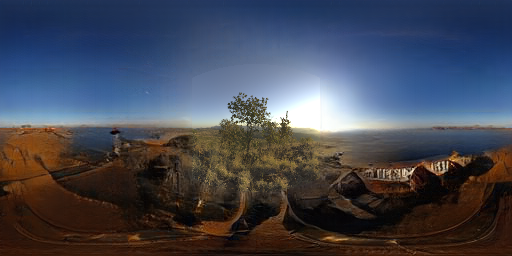}
    % & \includegraphics[width=\mywidth]{figs/qualitative_outdoor/pano/evrlight/lenong_2_8k_00.png}
    & \includegraphics[width=\mywidth]{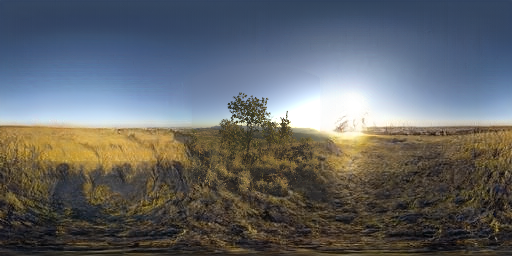}
    \\
    \includegraphics[width=\mywidth]{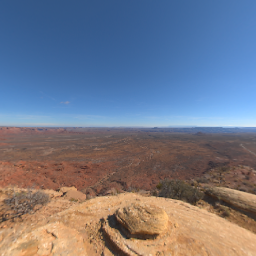}
    & \includegraphics[width=\mywidth]{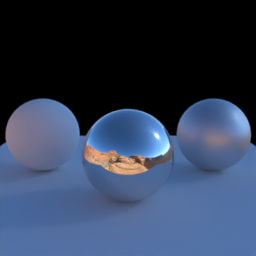}
    & \includegraphics[width=\mywidth]{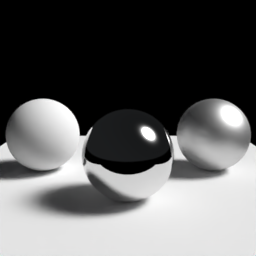}
    % & \includegraphics[width=\mywidth]{figs/qualitative_outdoor/render/immersegan/Serpentine_Valley_3k_00.png}
    & \includegraphics[width=\mywidth]{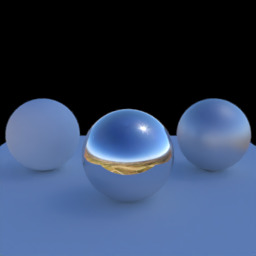}
    % & \includegraphics[width=\mywidth]{figs/qualitative_outdoor/render/everlight/Serpentine_Valley_3k_00.png}
    & \includegraphics[width=\mywidth]{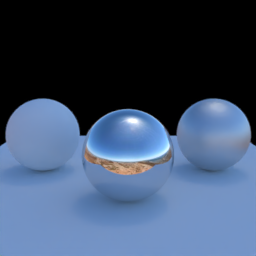}
    \\
    & \includegraphics[width=\mywidth]{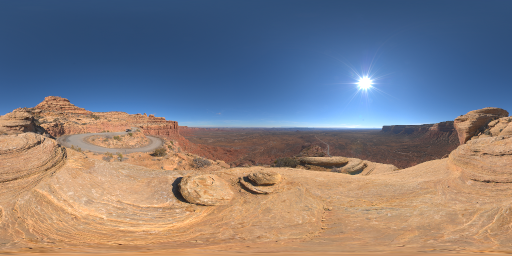}
    & \includegraphics[width=\mywidth]{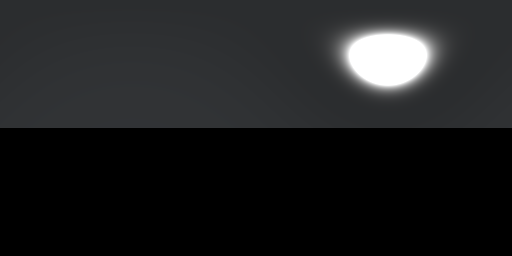}
    % & \includegraphics[width=\mywidth]{figs/qualitative_outdoor/pano/immersegan/Serpentine_Valley_3k_00.png}
    & \includegraphics[width=\mywidth]{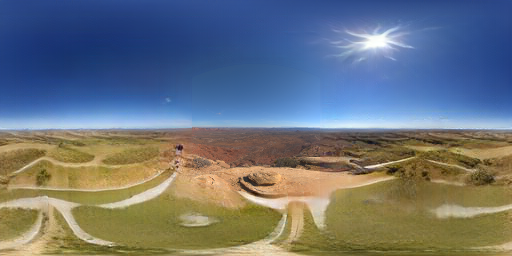}
    % & \includegraphics[width=\mywidth]{figs/qualitative_outdoor/pano/evrlight/Serpentine_Valley_3k_00.png}
    & \includegraphics[width=\mywidth]{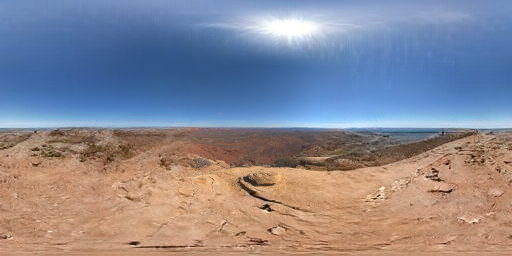}
    \\
    \includegraphics[width=\mywidth]{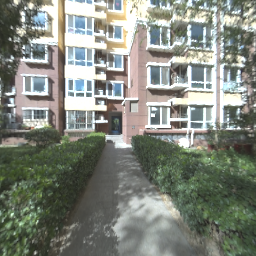}
    & \includegraphics[width=\mywidth]{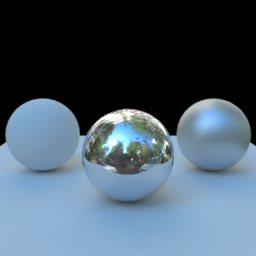}
    & \includegraphics[width=\mywidth]{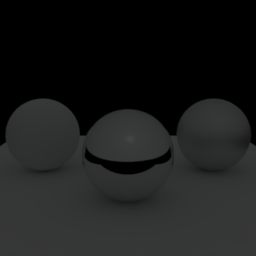}
    % & \includegraphics[width=\mywidth]{figs/qualitative_outdoor/render/immersegan/00167_00.png}
    & \includegraphics[width=\mywidth]{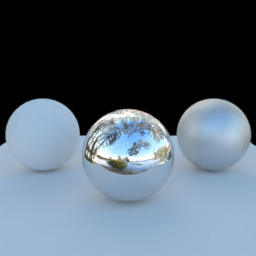}
    % & \includegraphics[width=\mywidth]{figs/qualitative_outdoor/render/everlight/00167_00.png}
    & \includegraphics[width=\mywidth]{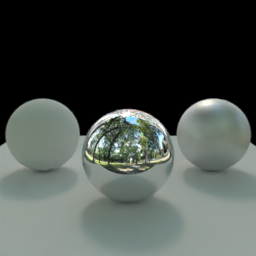}
    \\
    & \includegraphics[width=\mywidth]{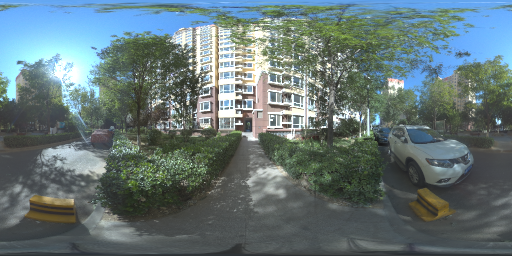}
    & \includegraphics[width=\mywidth]{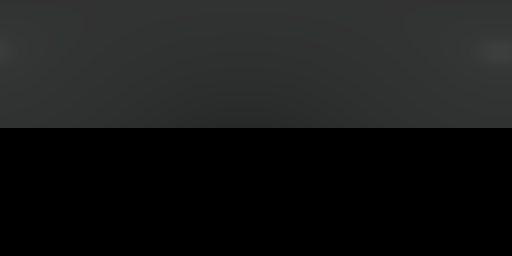}
    % & \includegraphics[width=\mywidth]{figs/qualitative_outdoor/pano/immersegan/00167_00.png}
    & \includegraphics[width=\mywidth]{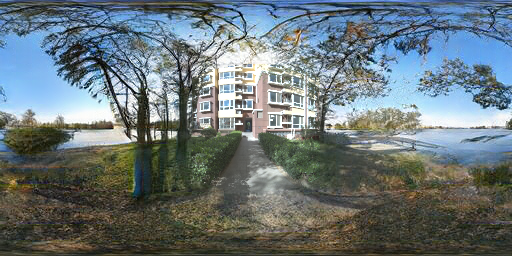}
    % & \includegraphics[width=\mywidth]{figs/qualitative_outdoor/pano/evrlight/00167_00.png}
    & \includegraphics[width=\mywidth]{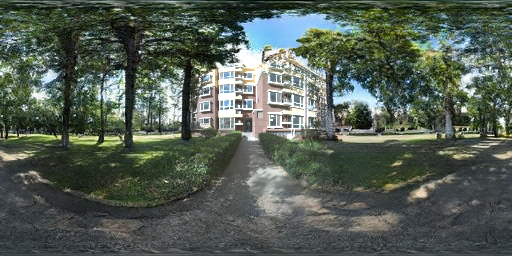}
    \\
    \end{tabular}}
    \caption[]{Qualitative rendering results for outdoor scenes. We compare against, from left to right: Zhang~\etal~2019~\cite{zhang2019all}, and ImmerseGAN~\cite{karimi2022guided}. Our results are shown in the last column. All methods with editable lighting capabilities are indicated with *. From a given input crop (left), we show a render of a scene composed of three spheres (diffuse, mirror, glossy) on a diffuse ground plane (first row), and the corresponding estimated lighting (in equirectangular format) below.}
    \label{fig:qual-outdoor}
\end{figure}

%% file: tables/qualitative_edit.tex
\begin{figure} [t]
    \centering
    \renewcommand{\tabcolsep}{1pt}
    \newcommand{\mywidth}{0.25\linewidth}
    \resizebox{\linewidth}{!}{% take the[] entire width, but still find a good per-image width otherwise text gets compressed
    \begin{tabular}{ccccc}
    Input & GT & Output & No Lights & Add \\ 
    \includegraphics[width=\mywidth]{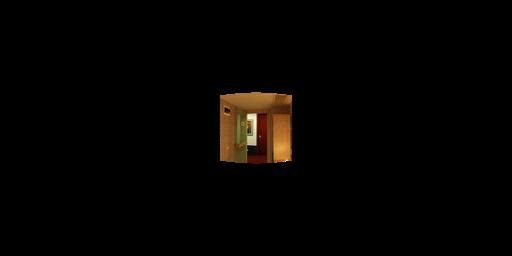}
    & \includegraphics[width=\mywidth]{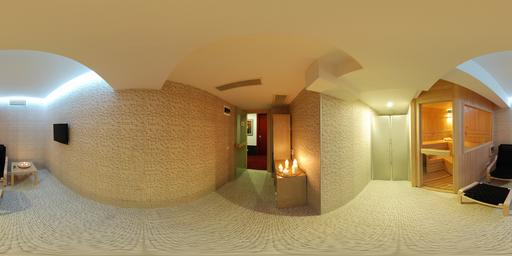}
    % & \includegraphics[width=\mywidth]{figs/editing/85815.jpg_pred.jpg}
    & \includegraphics[width=\mywidth]{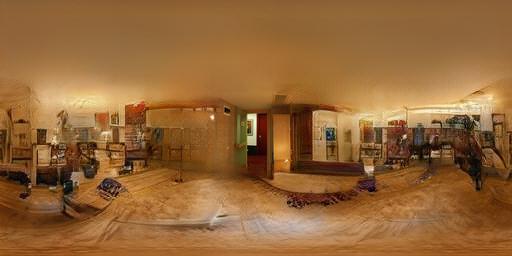}
    % & \includegraphics[width=\mywidth]{figs/editing/85815.jpg_zero.jpg}
    & \includegraphics[width=\mywidth]{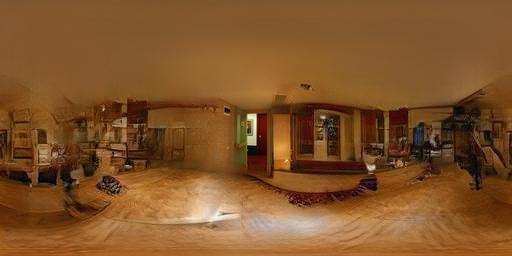}
    % & \includegraphics[width=\mywidth]{figs/editing/85815.jpg_11596_SG.blosc.jpg}
    % & \includegraphics[width=\mywidth]{figs/editing/85815.jpg_556194_SG.blosc.jpg}
    & \includegraphics[width=\mywidth]{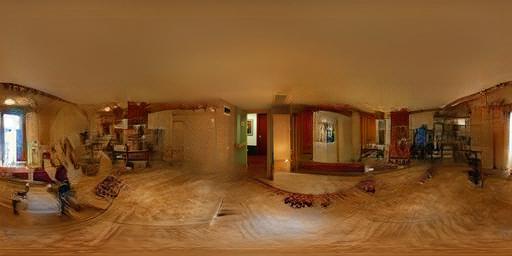}
    \\
    & \includegraphics[width=\mywidth]{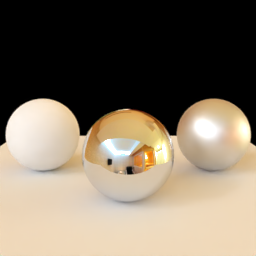}
    % & \includegraphics[width=\mywidth]{figs/editing/render/85815.jpg_pred.jpg}
    & \includegraphics[width=\mywidth]{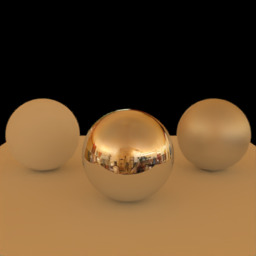}
    % & \includegraphics[width=\mywidth]{figs/editing/render/85815.jpg_zero.jpg}
    & \includegraphics[width=\mywidth]{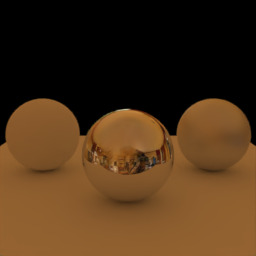}
    % & \includegraphics[width=\mywidth]{figs/editing/render/85815.jpg_11596_SG.blosc.jpg}
    % & \includegraphics[width=\mywidth]{figs/editing/render/85815.jpg_556194_SG.blosc.jpg}
    & \includegraphics[width=\mywidth]{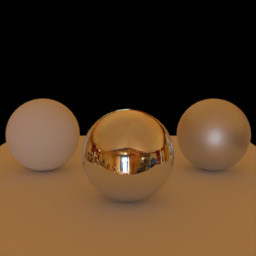}
    \\

    Input & GT & Output & No Sun & New Sun\\ 
    \includegraphics[width=\mywidth]{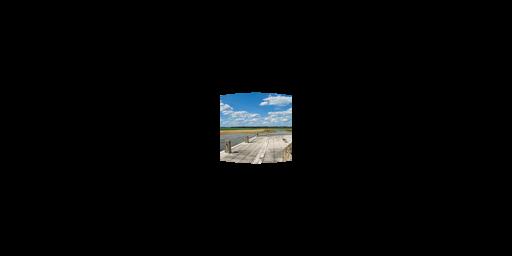}
    & \includegraphics[width=\mywidth]{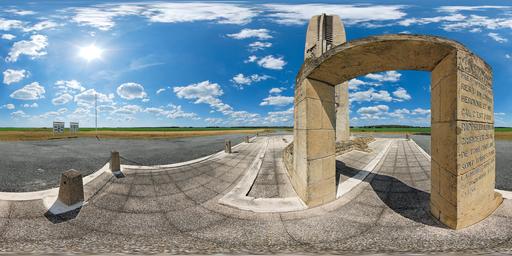}
    % & \includegraphics[width=\mywidth]{figs/editing/153703.jpg_pred.jpg}
    & \includegraphics[width=\mywidth]{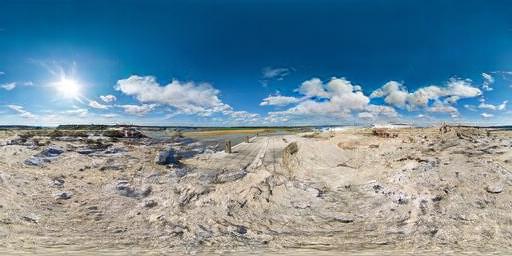}
    % & \includegraphics[width=\mywidth]{figs/editing/153703.jpg_zero.jpg}
    & \includegraphics[width=\mywidth]{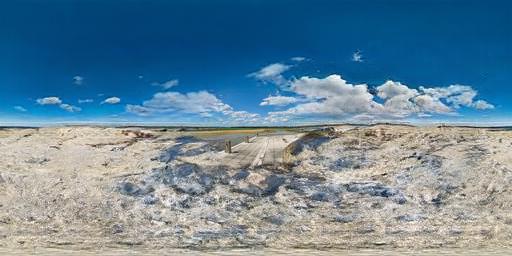}
    % & \includegraphics[width=\mywidth]{figs/editing/153703.jpg_260921_SG.blosc.jpg}
    & \includegraphics[width=\mywidth]{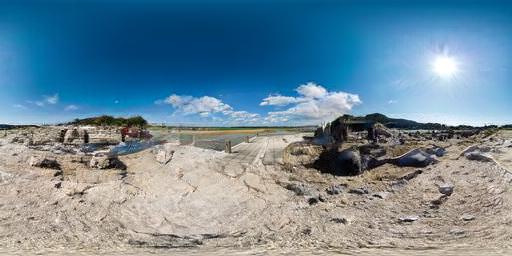}
    \\
    & \includegraphics[width=\mywidth]{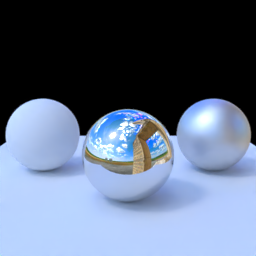}
    % & \includegraphics[width=\mywidth]{figs/editing/render/153703.jpg_pred.jpg}
    & \includegraphics[width=\mywidth]{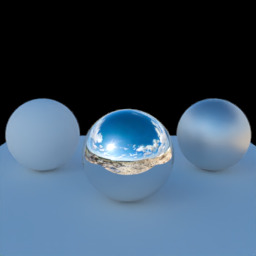}
    % & \includegraphics[width=\mywidth]{figs/editing/render/153703.jpg_zero.jpg}
    & \includegraphics[width=\mywidth]{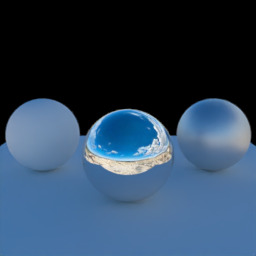}
    % & \includegraphics[width=\mywidth]{figs/editing/render/153703.jpg_260921_SG.blosc.jpg}
    & \includegraphics[width=\mywidth]{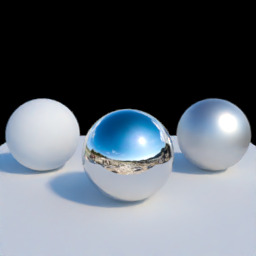}
    \\
    \end{tabular}}
    \caption{Qualitative light editing results. For each group of results (top: indoor, bottom: outdoor), we show the input (image projected on equirectangular representation), ground truth, automatic estimate (``output''), removing the lights, and adding other lights. The first row shows the environment map, and second row the corresponding rendering. }
    \label{fig:qual-editing}
\end{figure}

%% file: tables/qualitative_IMGxSG.tex
\begin{figure*} [t]
    \centering
    \renewcommand{\tabcolsep}{1pt}
    \newcommand{\mywidth}{0.25\linewidth}
    \resizebox{\linewidth}{!}{% take the[] entire width, but still find a good per-image width otherwise text gets compressed
    \begin{tabular}{ccccc}
    & \includegraphics[width=\mywidth]{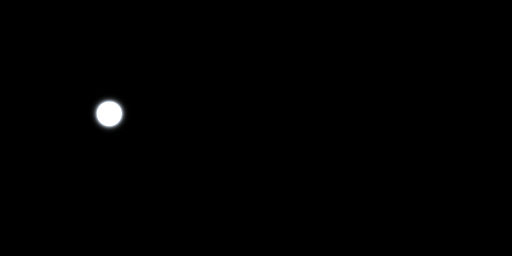}
    & \includegraphics[width=\mywidth]{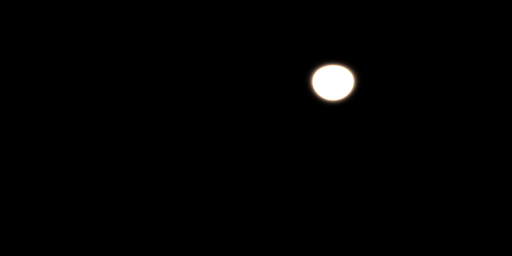}
    & \includegraphics[width=\mywidth]{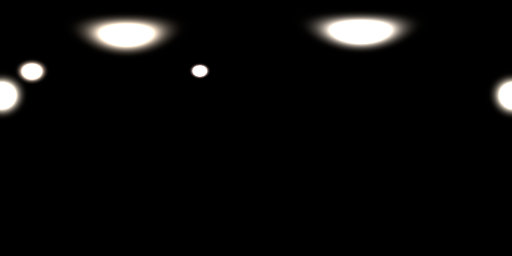}
    & \includegraphics[width=\mywidth]{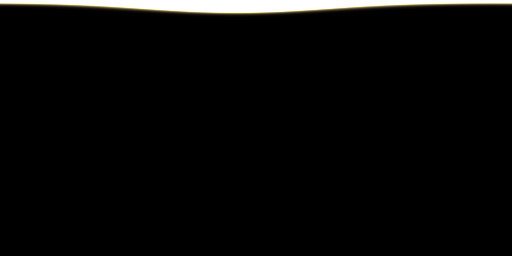}
    \\
    % \includegraphics[width=\mywidth]{figs/IMGxSG/17378_masked.jpg}
    % & \includegraphics[width=\mywidth]{figs/IMGxSG/17378_17378.jpg}
    % & \includegraphics[width=\mywidth]{figs/IMGxSG/17378_32706.jpg}
    % & \includegraphics[width=\mywidth]{figs/IMGxSG/17378_148928.jpg}
    % & \includegraphics[width=\mywidth]{figs/IMGxSG/17378_54879.jpg}
    % \\
    \includegraphics[width=\mywidth]{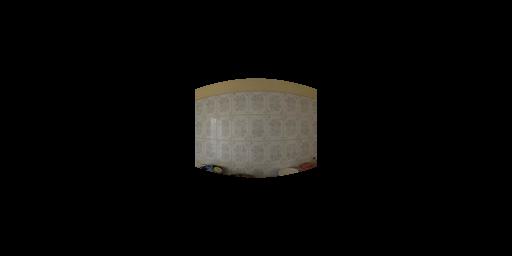}
    & \includegraphics[width=\mywidth]{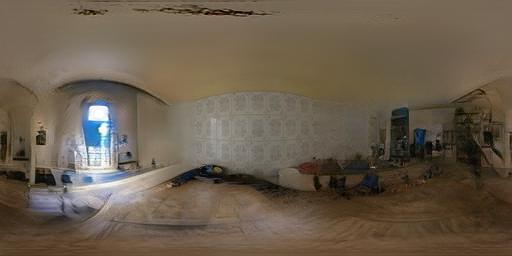}
    & \includegraphics[width=\mywidth]{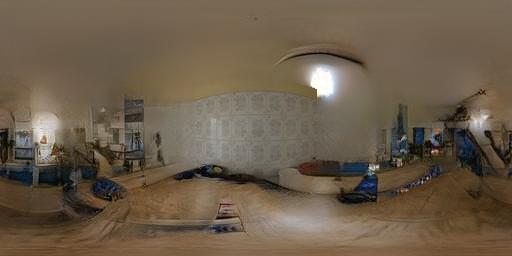}
    & \includegraphics[width=\mywidth]{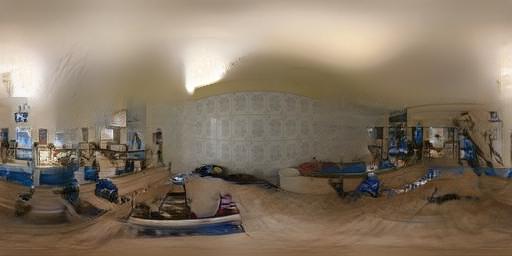}
    & \includegraphics[width=\mywidth]{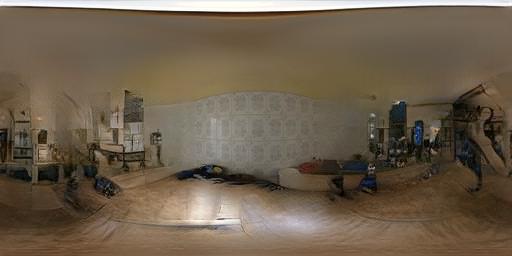}
    \\
    % & \includegraphics[width=\mywidth]{figs/imgxsg_render/17378_17378.jpg}
    % & \includegraphics[width=\mywidth]{figs/imgxsg_render/17378_32706.jpg}
    % & \includegraphics[width=\mywidth]{figs/imgxsg_render/17378_148928.jpg}
    % & \includegraphics[width=\mywidth]{figs/imgxsg_render/17378_54879.jpg}
    % \\
    & \includegraphics[width=\mywidth]{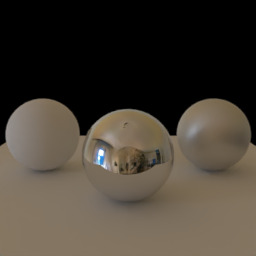}
    & \includegraphics[width=\mywidth]{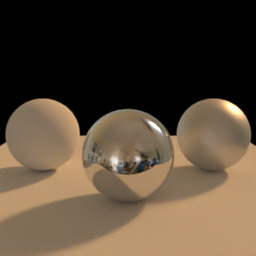}
    & \includegraphics[width=\mywidth]{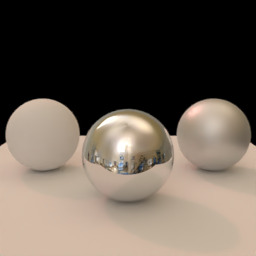}
    & \includegraphics[width=\mywidth]{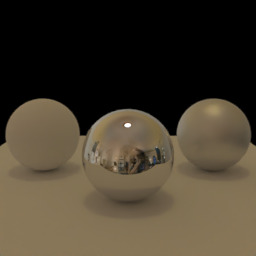}
    \\
    
    % \includegraphics[width=\mywidth]{figs/IMGxSG/32706_masked.jpg}
    % & \includegraphics[width=\mywidth]{figs/IMGxSG/32706_17378.jpg}
    % & \includegraphics[width=\mywidth]{figs/IMGxSG/32706_32706.jpg}
    % & \includegraphics[width=\mywidth]{figs/IMGxSG/32706_148928.jpg}
    % & \includegraphics[width=\mywidth]{figs/IMGxSG/32706_54879.jpg}
    % \\
     \includegraphics[width=\mywidth]{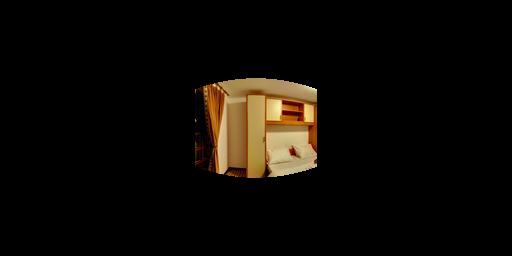}
    & \includegraphics[width=\mywidth]{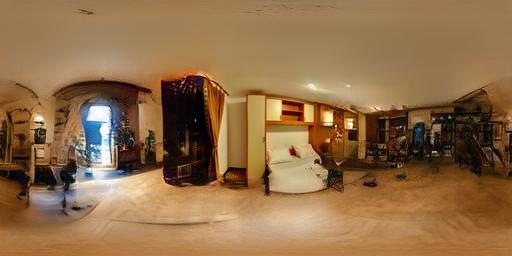}
    & \includegraphics[width=\mywidth]{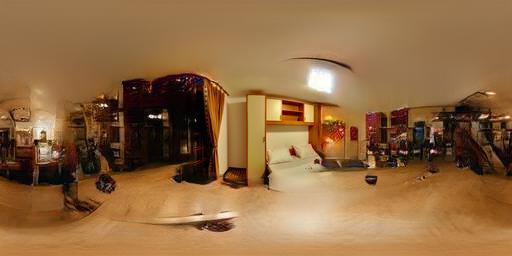}
    & \includegraphics[width=\mywidth]{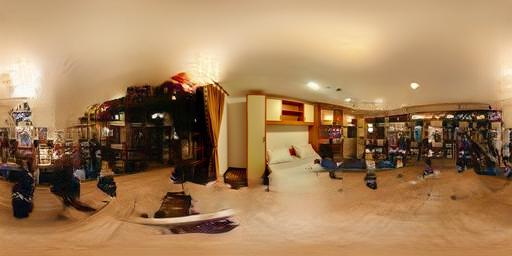}
    & \includegraphics[width=\mywidth]{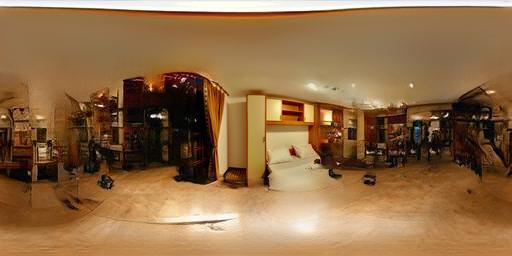}
    \\
    % & \includegraphics[width=\mywidth]{figs/imgxsg_render/32706_17378.jpg}
    % & \includegraphics[width=\mywidth]{figs/imgxsg_render/32706_32706.jpg}
    % & \includegraphics[width=\mywidth]{figs/imgxsg_render/32706_148928.jpg}
    % & \includegraphics[width=\mywidth]{figs/imgxsg_render/32706_54879.jpg}
    % \\
    & \includegraphics[width=\mywidth]{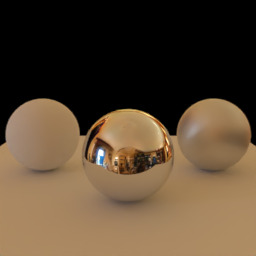}
    & \includegraphics[width=\mywidth]{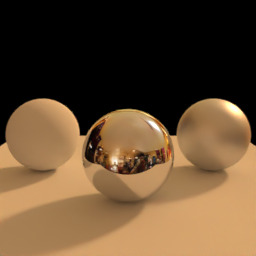}
    & \includegraphics[width=\mywidth]{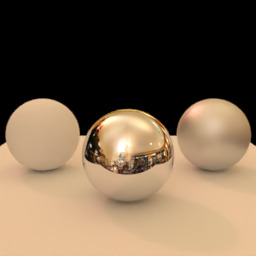}
    & \includegraphics[width=\mywidth]{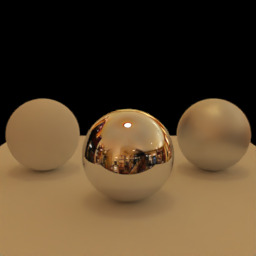}
    \\
    % \includegraphics[width=\mywidth]{figs/IMGxSG/59041_masked.jpg}
    % & \includegraphics[width=\mywidth]{figs/IMGxSG/102999_17378.jpg}
    % & \includegraphics[width=\mywidth]{figs/IMGxSG/102999_32706.jpg}
    % & \includegraphics[width=\mywidth]{figs/IMGxSG/102999_148928.jpg}
    % & \includegraphics[width=\mywidth]{figs/IMGxSG/102999_54879.jpg}
    % \\
    \includegraphics[width=\mywidth]{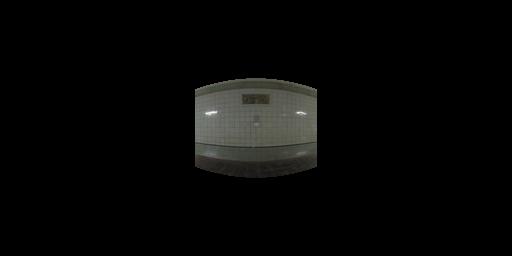}
    & \includegraphics[width=\mywidth]{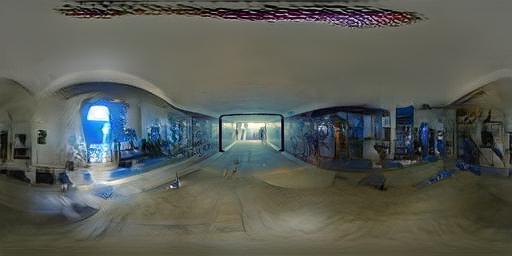}
    & \includegraphics[width=\mywidth]{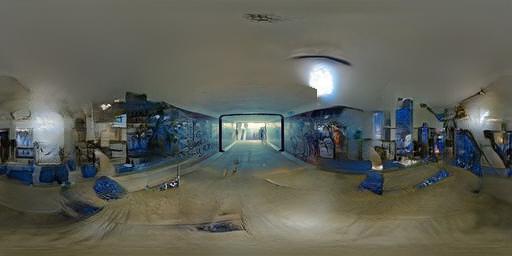}
    & \includegraphics[width=\mywidth]{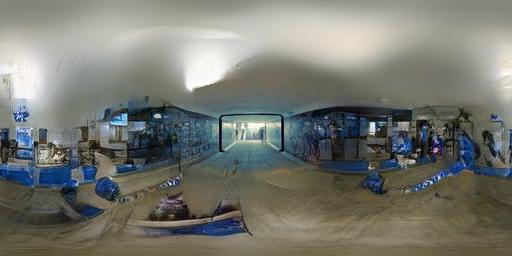}
    & \includegraphics[width=\mywidth]{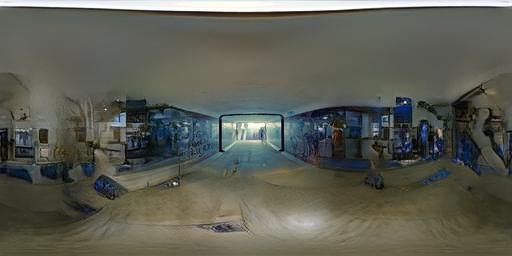}
    \\
    % & \includegraphics[width=\mywidth]{figs/imgxsg_render/102999_17378.jpg}
    % & \includegraphics[width=\mywidth]{figs/imgxsg_render/102999_32706.jpg}
    % & \includegraphics[width=\mywidth]{figs/imgxsg_render/102999_148928.jpg}
    % & \includegraphics[width=\mywidth]{figs/imgxsg_render/102999_54879.jpg}
    & \includegraphics[width=\mywidth]{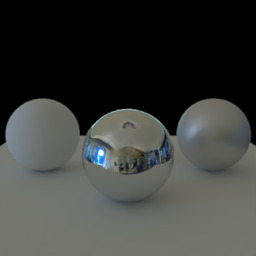}
    & \includegraphics[width=\mywidth]{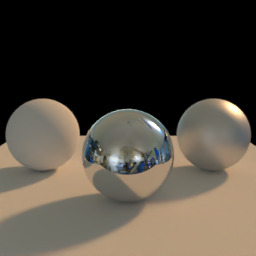}
    & \includegraphics[width=\mywidth]{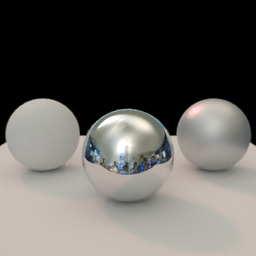}
    & \includegraphics[width=\mywidth]{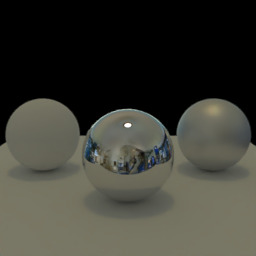}
    \end{tabular}
    \begin{tabular}{ccccc}
    & \includegraphics[width=\mywidth]{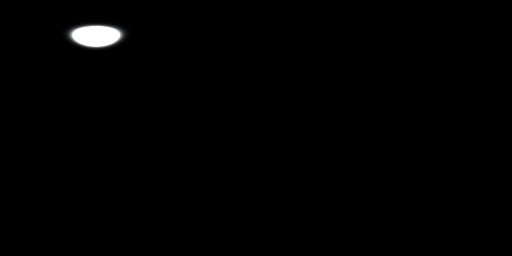}
    & \includegraphics[width=\mywidth]{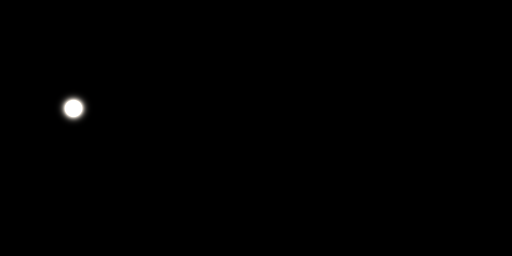}
    & \includegraphics[width=\mywidth]{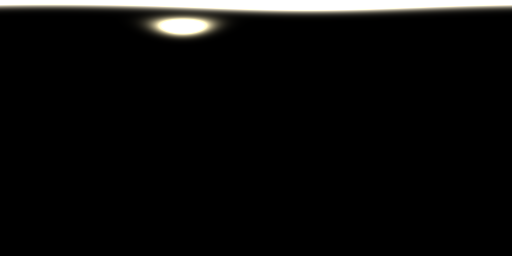}
    & \includegraphics[width=\mywidth]{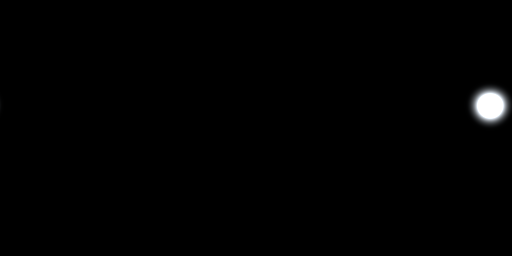}
    \\
    % \includegraphics[width=\mywidth]{figs/IMGxSG/35860_masked.jpg}
    % & \includegraphics[width=\mywidth]{figs/IMGxSG_outdoor/35860_35860.jpg}
    % & \includegraphics[width=\mywidth]{figs/IMGxSG_outdoor/35860_56348.jpg}
    % & \includegraphics[width=\mywidth]{figs/IMGxSG_outdoor/35860_100588.jpg}
    % & \includegraphics[width=\mywidth]{figs/IMGxSG_outdoor/35860_620723.jpg}
    % \\
    \includegraphics[width=\mywidth]{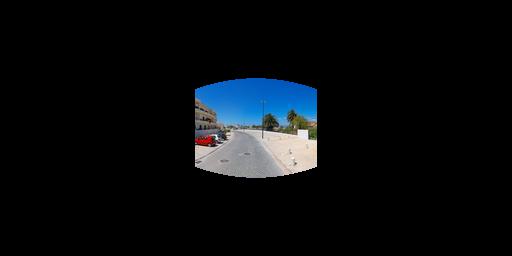}
    & \includegraphics[width=\mywidth]{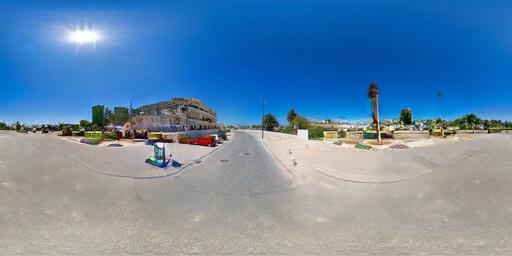}
    & \includegraphics[width=\mywidth]{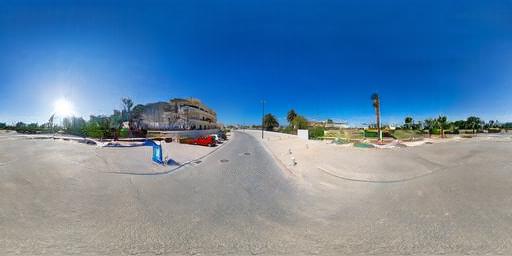}
    & \includegraphics[width=\mywidth]{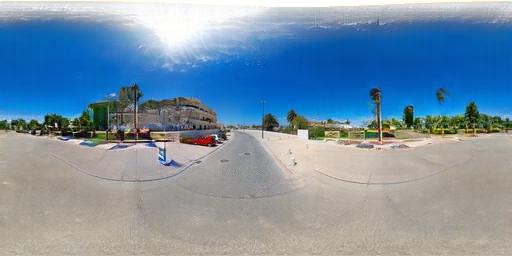}
    & \includegraphics[width=\mywidth]{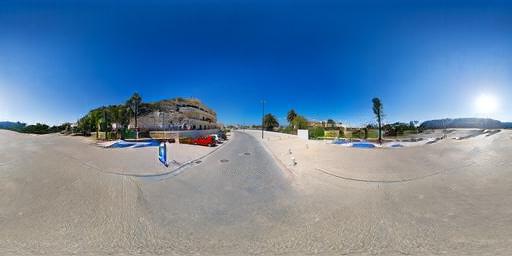}
    \\
    % & \includegraphics[width=\mywidth]{figs/imgxsg_render/35860_35860.jpg}
    % & \includegraphics[width=\mywidth]{figs/imgxsg_render/35860_56348.jpg}
    % & \includegraphics[width=\mywidth]{figs/imgxsg_render/35860_100588.jpg}
    % & \includegraphics[width=\mywidth]{figs/imgxsg_render/35860_620723.jpg}
    % \\
    & \includegraphics[width=\mywidth]{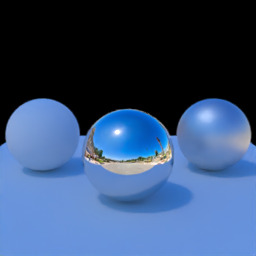}
    & \includegraphics[width=\mywidth]{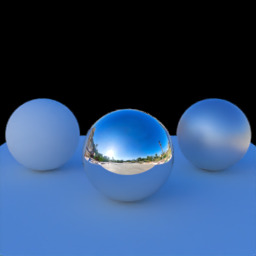}
    & \includegraphics[width=\mywidth]{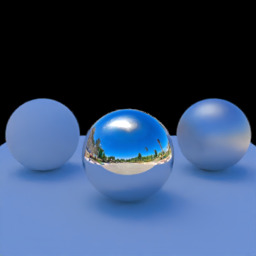}
    & \includegraphics[width=\mywidth]{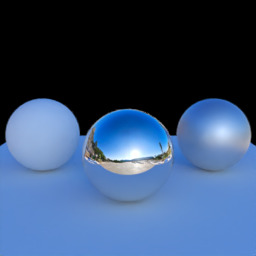}
    \\
    % \includegraphics[width=\mywidth]{figs/IMGxSG/56348_masked.jpg}
    % & \includegraphics[width=\mywidth]{figs/IMGxSG_outdoor/56348_35860.jpg}
    % & \includegraphics[width=\mywidth]{figs/IMGxSG_outdoor/56348_56348.jpg}
    % & \includegraphics[width=\mywidth]{figs/IMGxSG_outdoor/56348_100588.jpg}
    % & \includegraphics[width=\mywidth]{figs/IMGxSG_outdoor/56348_620723.jpg}
    % \\
    \includegraphics[width=\mywidth]{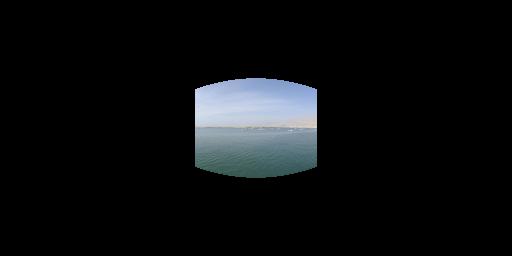}
    & \includegraphics[width=\mywidth]{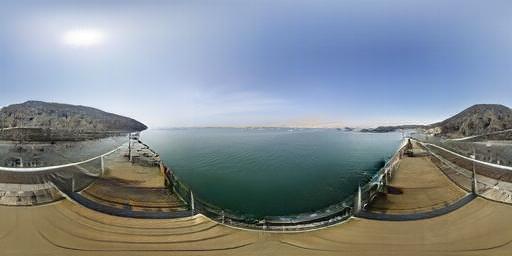}
    & \includegraphics[width=\mywidth]{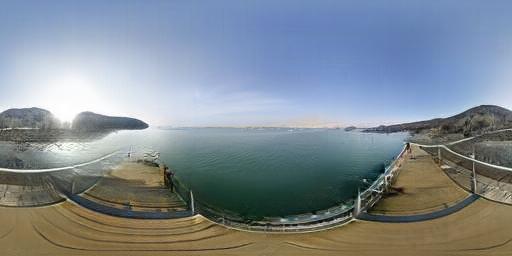}
    & \includegraphics[width=\mywidth]{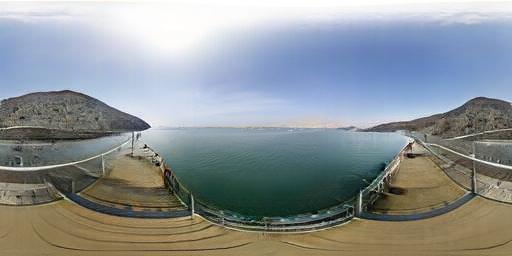}
    & \includegraphics[width=\mywidth]{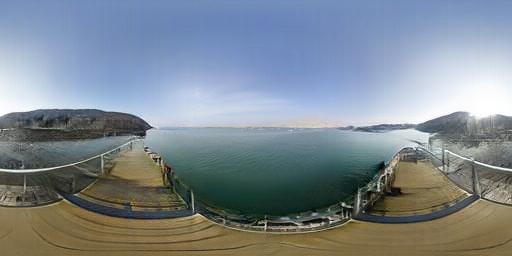}
    \\
    % & \includegraphics[width=\mywidth]{figs/imgxsg_render/56348_35860.jpg}
    % & \includegraphics[width=\mywidth]{figs/imgxsg_render/56348_56348.jpg}
    % & \includegraphics[width=\mywidth]{figs/imgxsg_render/56348_100588.jpg}
    % & \includegraphics[width=\mywidth]{figs/imgxsg_render/56348_620723.jpg}
    % \\
    & \includegraphics[width=\mywidth]{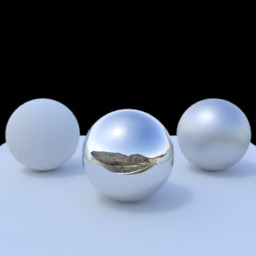}
    & \includegraphics[width=\mywidth]{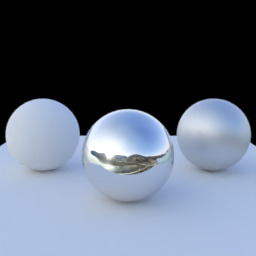}
    & \includegraphics[width=\mywidth]{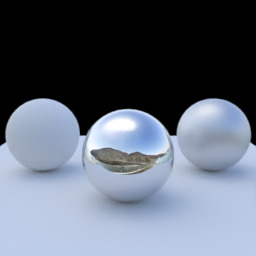}
    & \includegraphics[width=\mywidth]{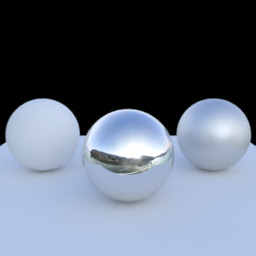}
    \\
    % \includegraphics[width=\mywidth]{figs/IMGxSG/100588_masked.jpg}
    % & \includegraphics[width=\mywidth]{figs/IMGxSG_outdoor/100588_35860.jpg}
    % & \includegraphics[width=\mywidth]{figs/IMGxSG_outdoor/100588_56348.jpg}
    % & \includegraphics[width=\mywidth]{figs/IMGxSG_outdoor/100588_100588.jpg}
    % & \includegraphics[width=\mywidth]{figs/IMGxSG_outdoor/100588_620723.jpg}
    % \\
    \includegraphics[width=\mywidth]{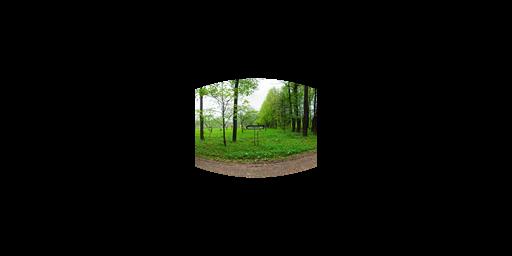}
    & \includegraphics[width=\mywidth]{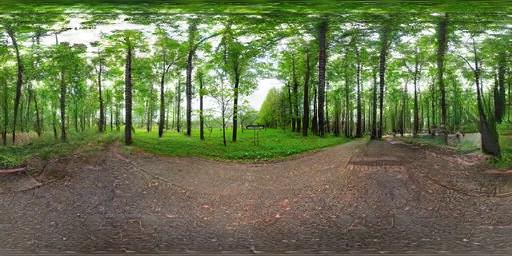}
    & \includegraphics[width=\mywidth]{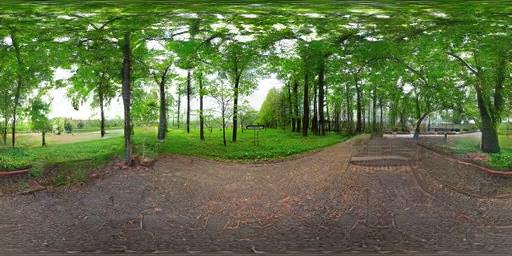}
    & \includegraphics[width=\mywidth]{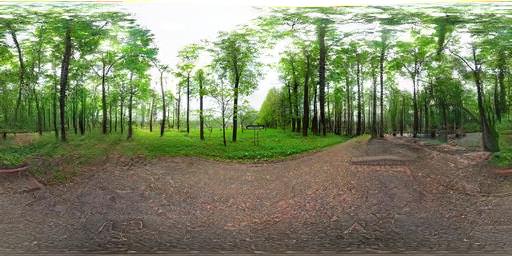}
    & \includegraphics[width=\mywidth]{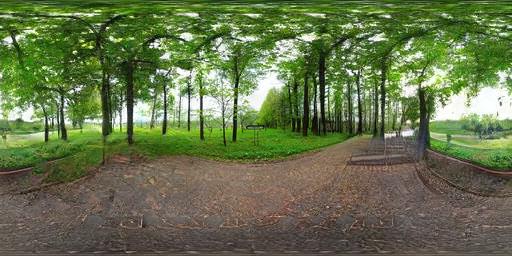}
    \\
    % & \includegraphics[width=\mywidth]{figs/imgxsg_render/100588_35860.jpg}
    % & \includegraphics[width=\mywidth]{figs/imgxsg_render/100588_56348.jpg}
    % & \includegraphics[width=\mywidth]{figs/imgxsg_render/100588_100588.jpg}
    % & \includegraphics[width=\mywidth]{figs/imgxsg_render/100588_620723.jpg}
    % \\
    & \includegraphics[width=\mywidth]{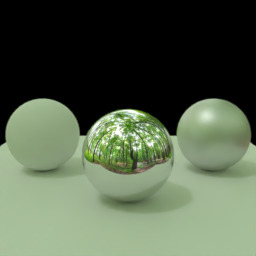}
    & \includegraphics[width=\mywidth]{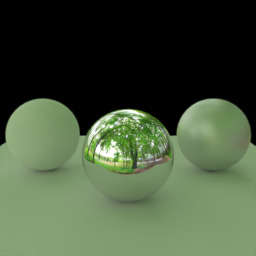}
    & \includegraphics[width=\mywidth]{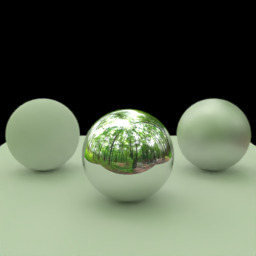}
    & \includegraphics[width=\mywidth]{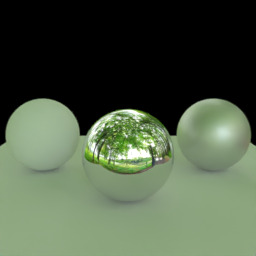}
    \\
    \end{tabular}}
    \caption{Qualitative light editing results. Each group of results (left: indoor, right: outdoor) show, on the first column: 3 different input images projected in equirectangular representation; and on the first row: 4 different lighting configurations. The $3 \times 4$ image matrix correspond to the environment map predicted by \thename for each input/lighting combination. Note how the network learned to realistically blend the bright light sources with their environments. For example, the virtual light sources create reflections on the ground (top) and water (below). }
    \label{fig:qual-imgxsg}
\end{figure*}

%% file: 5_discussion.tex
\section{Discussion}

We propose a method for estimating lighting from both indoor and outdoor environments as editable HDR \ang{360} panoramas from regular images. Doing so effectively bridges the gap in currently available methods in the literature, where most methods are specifically designed for either indoors or outdoors, offer limited editing capabilities, or employ simplified, strictly parametric lighting models. Our approach enables easy user editing of several key lighting parameters, including adding and removing lights, and editing their intensity, color, and direction, while simultaneously providing high-quality texture within the HDR panorama to bring reflections to life when performing virtual object insertion. Notwithstanding being generic to in- and outdoor environments, our method remains quantitatively competitive with domain-specific techniques. 
We are hopeful our ideas help reconcile the current schism in indoor and outdoor lighting estimation in the literature. 

%\paragraph{Limitations} 
Despite providing either state-of-the-art or competitive results on several aspects of lighting and panorama texture generation, our method bears some limitations that we are hopeful future work will lift. First, the choice of spherical gaussians for HDR light sources is a trade-off modeling adequately the majority of light sources present in scenes in-the-wild. However, it does not visually represent many light sources like window panes, or accurately model very bright sources like the sun~\cite{lalonde2014lighting}. Second, while the part of our lighting representation that influences shading is flexible and editable, the panorama texture is currently not editable. We believe that integrating a guiding system for the generator, such as in \cite{karimi2022guided}, would bring back editability to the texture part of our representation. Lastly, our light predictor $\mathcal{L}$ is trained on proxy ground-truth data predicted by a network, leading to limited accuracy in light position and intensity. Capturing a large-scale dataset of real ground-truth HDR environments would likely help improve the accuracy.

%% file: main.bbl
\begin{thebibliography}{10}\itemsep=-1pt

\bibitem{akimoto2019360}
Naofumi Akimoto, Seito Kasai, Masaki Hayashi, and Yoshimitsu Aoki.
\newblock 360-degree image completion by two-stage conditional {GAN}s.
\newblock In {\em ICIP}, 2019.

\bibitem{akimoto2022diverse}
Naofumi Akimoto, Yuhi Matsuo, and Yoshimitsu Aoki.
\newblock Diverse plausible 360-degree image outpainting for efficient 3dcg
  background creation.
\newblock In {\em CVPR}, 2022.

\bibitem{barnes2009patchmatch}
Connelly Barnes, Eli Shechtman, Adam Finkelstein, and Dan~B Goldman.
\newblock Patchmatch: A randomized correspondence algorithm for structural
  image editing.
\newblock {\em ACM TOG}, 28(3):24, 2009.

\bibitem{barron2014shape}
Jonathan~T Barron and Jitendra Malik.
\newblock Shape, illumination, and reflectance from shading.
\newblock {\em IEEE TPAMI}, 37(8):1670--1687, 2014.

\bibitem{cheng2018shlight}
Dachuan Cheng, Jian Shi, Yanyun Chen, Xiaoming Deng, and Xiaopeng. Zhang.
\newblock Learning scene illumination by pairwise photos from rear and front
  mobile cameras.
\newblock {\em Computer Graphics Forum}, 37(7):213--221, 2018.

\bibitem{karimi2022guided}
Mohammad Reza~Karimi Dastjerdi, Yannick Hold-Geoffroy, Jonathan Eisenman,
  Siavash Khodadadeh, and Jean-Fran\c{c}ois Lalonde.
\newblock Guided co-modulated {GAN} for \ang{360} field of view extrapolation.
\newblock In {\em 3DV}, 2022.

\bibitem{debevec1998rendering}
Paul Debevec.
\newblock Rendering synthetic objects into real scenes: Bridging traditional
  and image-based graphics with global illumination and high dynamic range
  photography.
\newblock In {\em Proceedings of the 25th Annual Conference on Computer
  Graphics and Interactive Techniques}, SIGGRAPH, pages 189--198, 1998.

\bibitem{efros2001image}
Alexei~A Efros and William~T Freeman.
\newblock Image quilting for texture synthesis and transfer.
\newblock In {\em Proceedings of the 28th annual conference on Computer
  graphics and interactive techniques}, pages 341--346, 2001.

\bibitem{efros1999texture}
Alexei~A Efros and Thomas~K Leung.
\newblock Texture synthesis by non-parametric sampling.
\newblock In {\em ICCV}, volume~2, pages 1033--1038. IEEE, 1999.

\bibitem{eilertsen2017hdr}
Gabriel Eilertsen, Joel Kronander, Gyorgy Denes, Rafa{\l}~K Mantiuk, and Jonas
  Unger.
\newblock Hdr image reconstruction from a single exposure using deep cnns.
\newblock {\em ACM TOG}, 36(6):1--15, 2017.

\bibitem{endo2017deep}
Yuki Endo, Yoshihiro Kanamori, and Jun Mitani.
\newblock Deep reverse tone mapping.
\newblock {\em ACM TOG}, 36(6):177--1, 2017.

\bibitem{gardner2019deep}
Marc-Andr{\'e} Gardner, Yannick Hold-Geoffroy, Kalyan Sunkavalli, Christian
  Gagn{\'e}, and Jean-Fran{\c{c}}ois Lalonde.
\newblock Deep parametric indoor lighting estimation.
\newblock In {\em ICCV}, 2019.

\bibitem{gardner2017learning}
Marc-Andr{\'e} Gardner, Kalyan Sunkavalli, Ersin Yumer, Xiaohui Shen, Emiliano
  Gambaretto, Christian Gagn{\'e}, and Jean-Fran{\c{c}}ois Lalonde.
\newblock Learning to predict indoor illumination from a single image.
\newblock {\em ACM TOG}, 9(4), 2017.

\bibitem{garon2019fast}
Mathieu Garon, Kalyan Sunkavalli, Sunil Hadap, Nathan Carr, and
  Jean-Fran{\c{c}}ois Lalonde.
\newblock Fast spatially-varying indoor lighting estimation.
\newblock In {\em CVPR}, 2019.

\bibitem{griffiths2022outcast}
David Griffiths, Tobias Ritschel, and Julien Philip.
\newblock {OutCast}: Outdoor single-image relighting with cast shadows.
\newblock {\em Computer Graphics Forum}, 41(2):179--193, 2022.

\bibitem{guillemot2013image}
Christine Guillemot and Olivier Le~Meur.
\newblock Image inpainting: Overview and recent advances.
\newblock {\em IEEE signal processing magazine}, 31(1):127--144, 2013.

\bibitem{hara2021spherical}
Takayuki Hara, Yusuke Mukuta, and Tatsuya Harada.
\newblock Spherical image generation from a single image by considering scene
  symmetry.
\newblock In {\em AAAI}, 2021.

\bibitem{Hartley2004}
R.~I. Hartley and A. Zisserman.
\newblock {\em Multiple View Geometry in Computer Vision}.
\newblock Cambridge University Press, second edition, 2004.

\bibitem{heusel2017gans}
Martin Heusel, Hubert Ramsauer, Thomas Unterthiner, Bernhard Nessler, and Sepp
  Hochreiter.
\newblock Gans trained by a two time-scale update rule converge to a local nash
  equilibrium.
\newblock In {\em NeurIPS}, 2017.

\bibitem{holdgeoffroy2019deep}
Yannick Hold-Geoffroy, Akshaya Athawale, and Jean-Fran\c{c}ois Lalonde.
\newblock Deep sky modeling for single image outdoor lighting estimation.
\newblock In {\em CVPR}, 2019.

\bibitem{hold2017deep}
Yannick Hold-Geoffroy, Kalyan Sunkavalli, Sunil Hadap, Emiliano Gambaretto, and
  Jean-Fran{\c{c}}ois Lalonde.
\newblock Deep outdoor illumination estimation.
\newblock In {\em CVPR}, 2017.

\bibitem{isola2017image}
Phillip Isola, Jun-Yan Zhu, Tinghui Zhou, and Alexei~A Efros.
\newblock Image-to-image translation with conditional adversarial networks.
\newblock In {\em CVPR}, 2017.

\bibitem{karras2019style}
Tero Karras, Samuli Laine, and Timo Aila.
\newblock A style-based generator architecture for generative adversarial
  networks.
\newblock In {\em CVPR}, 2019.

\bibitem{kulkarni2022360fusionnerf}
Shreyas Kulkarni, Peng Yin, and Sebastian Scherer.
\newblock 360fusionnerf: Panoramic neural radiance fields with joint guidance.
\newblock {\em arXiv preprint arXiv:2209.14265}, 2022.

\bibitem{skydb}
Jean-Fran\c{c}ois Lalonde, Louis-Philippe Asselin, Julien Becirovski, Yannick
  Hold-Geoffroy, Mathieu Garon, Marc-Andr\'{e} Gardner, and Jinsong Zhang.
\newblock The {Laval} {HDR} sky database, 2016.

\bibitem{lalonde2012estimating}
Jean-Fran{\c{c}}ois Lalonde, Alexei~A Efros, and Srinivasa~G Narasimhan.
\newblock Estimating the natural illumination conditions from a single outdoor
  image.
\newblock {\em IJCV}, 98(2):123--145, 2012.

\bibitem{lalonde2014lighting}
Jean-Fran{\c{c}}ois Lalonde and Iain Matthews.
\newblock Lighting estimation in outdoor image collections.
\newblock In {\em 3DV}, 2014.

\bibitem{lalonde2010sun}
Jean-Fran{\c{c}}ois Lalonde, Srinivasa~G Narasimhan, and Alexei~A Efros.
\newblock What do the sun and the sky tell us about the camera?
\newblock {\em IJCV}, 88:24--51, 2010.

\bibitem{legendre2019deeplight}
Chloe LeGendre, Wan-Chun Ma, Graham Fyffe, John Flynn, Laurent Charbonnel, Jay
  Busch, and Paul Debevec.
\newblock Deeplight: Learning illumination for unconstrained mobile mixed
  reality.
\newblock In {\em CVPR}, 2019.

\bibitem{li2020inverse}
Zhengqin Li, Mohammad Shafiei, Ravi Ramamoorthi, Kalyan Sunkavalli, and
  Manmohan Chandraker.
\newblock Inverse rendering for complex indoor scenes: Shape, spatially-varying
  lighting and {SVBRDF} from a single image.
\newblock In {\em CVPR}, 2020.

\bibitem{li2022physically}
Zhengqin Li, Jia Shi, Sai Bi, Rui Zhu, Kalyan Sunkavalli, Milo{\v{s}}
  Ha{\v{s}}an, Zexiang Xu, Ravi Ramamoorthi, and Manmohan Chandraker.
\newblock Physically-based editing of indoor scene lighting from a single
  image.
\newblock In {\em ECCV}, 2022.

\bibitem{liu2020single}
Yu-Lun Liu, Wei-Sheng Lai, Yu-Sheng Chen, Yi-Lung Kao, Ming-Hsuan Yang, Yung-Yu
  Chuang, and Jia-Bin Huang.
\newblock Single-image hdr reconstruction by learning to reverse the camera
  pipeline.
\newblock In {\em CVPR}, 2020.

\bibitem{mandl2017learning}
David Mandl, Kwang~Moo Yi, Peter Mohr, Peter~M Roth, Pascal Fua, Vincent
  Lepetit, Dieter Schmalstieg, and Denis Kalkofen.
\newblock Learning lightprobes for mixed reality illumination.
\newblock In {\em IEEE Int. Symp. Mixed Aug. Reality}. IEEE, 2017.

\bibitem{marnerides2018expandnet}
Demetris Marnerides, Thomas Bashford-Rogers, Jonathan Hatchett, and Kurt
  Debattista.
\newblock Expandnet: A deep convolutional neural network for high dynamic range
  expansion from low dynamic range content.
\newblock {\em Computer Graphics Forum}, 37(2):37--49, 2018.

\bibitem{mildenhall2020nerf}
Ben Mildenhall, Pratul~P. Srinivasan, Matthew Tancik, Jonathan~T. Barron, Ravi
  Ramamoorthi, and Ren Ng.
\newblock Nerf: Representing scenes as neural radiance fields for view
  synthesis.
\newblock In {\em ECCV}, 2020.

\bibitem{paszke2019pytorch}
Adam Paszke, Sam Gross, Francisco Massa, Adam Lerer, James Bradbury, Gregory
  Chanan, Trevor Killeen, Zeming Lin, Natalia Gimelshein, Luca Antiga, et~al.
\newblock Pytorch: An imperative style, high-performance deep learning library.
\newblock {\em NeurIPS}, 2019.

\bibitem{radford2021learning}
Alec Radford, Jong~Wook Kim, Chris Hallacy, Aditya Ramesh, Gabriel Goh,
  Sandhini Agarwal, Girish Sastry, Amanda Askell, Pamela Mishkin, Jack Clark,
  et~al.
\newblock Learning transferable visual models from natural language
  supervision.
\newblock In {\em ICML}, 2021.

\bibitem{radford2015unsupervised}
Alec Radford, Luke Metz, and Soumith Chintala.
\newblock Unsupervised representation learning with deep convolutional
  generative adversarial networks.
\newblock {\em arXiv preprint arXiv:1511.06434}, 2015.

\bibitem{ramamoorthi2001efficient}
Ravi Ramamoorthi and Pat Hanrahan.
\newblock An efficient representation for irradiance environment maps.
\newblock In {\em Proceedings of the 28th annual conference on Computer
  graphics and interactive techniques}, pages 497--500, 2001.

\bibitem{sengupta2019neural}
Soumyadip Sengupta, Jinwei Gu, Kihwan Kim, Guilin Liu, David~W Jacobs, and Jan
  Kautz.
\newblock Neural inverse rendering of an indoor scene from a single image.
\newblock In {\em ICCV}, 2019.

\bibitem{shu2017neural}
Zhixin Shu, Ersin Yumer, Sunil Hadap, Kalyan Sunkavalli, Eli Shechtman, and
  Dimitris Samaras.
\newblock Neural face editing with intrinsic image disentangling.
\newblock In {\em CVPR}, 2017.

\bibitem{somanath2021hdr}
Gowri Somanath and Daniel Kurz.
\newblock {HDR} environment map estimation for real-time augmented reality.
\newblock In {\em CVPR}, 2021.

\bibitem{song2019neural}
Shuran Song and Thomas Funkhouser.
\newblock Neural illumination: Lighting prediction for indoor environments.
\newblock In {\em CVPR}, 2019.

\bibitem{srinivasan2020lighthouse}
Pratul~P Srinivasan, Ben Mildenhall, Matthew Tancik, Jonathan~T Barron, Richard
  Tucker, and Noah Snavely.
\newblock Lighthouse: Predicting lighting volumes for spatially-coherent
  illumination.
\newblock In {\em CVPR}, 2020.

\bibitem{tang2022estimating}
Jiajun Tang, Yongjie Zhu, Haoyu Wang, Jun~Hoong Chan, Si Li, and Boxin Shi.
\newblock Estimating spatially-varying lighting in urban scenes with
  disentangled representation.
\newblock In {\em ECCV}, 2022.

\bibitem{tewari2020state}
Ayush Tewari, Ohad Fried, Justus Thies, Vincent Sitzmann, Stephen Lombardi,
  Kalyan Sunkavalli, Ricardo Martin-Brualla, Tomas Simon, Jason Saragih,
  Matthias Nie{\ss}ner, et~al.
\newblock State of the art on neural rendering.
\newblock {\em Computer Graphics Forum}, 39(2):701--727, 2020.

\bibitem{wang2022stylelight}
Guangcong Wang, Yinuo Yang, Chen~Change Loy, and Ziwei Liu.
\newblock Stylelight: {HDR} panorama generation for lighting estimation and
  editing.
\newblock In {\em ECCV}, 2022.

\bibitem{wang2018high}
Ting-Chun Wang, Ming-Yu Liu, Jun-Yan Zhu, Andrew Tao, Jan Kautz, and Bryan
  Catanzaro.
\newblock High-resolution image synthesis and semantic manipulation with
  conditional {GANs}.
\newblock In {\em CVPR}, 2018.

\bibitem{wang2022neural}
Zian Wang, Wenzheng Chen, David Acuna, Jan Kautz, and Sanja Fidler.
\newblock Neural light field estimation for street scenes with differentiable
  virtual object insertion.
\newblock In {\em ECCV}, 2022.

\bibitem{wang2021learning}
Zian Wang, Jonah Philion, Sanja Fidler, and Jan Kautz.
\newblock Learning indoor inverse rendering with {3D} spatially-varying
  lighting.
\newblock In {\em ICCV}, 2021.

\bibitem{weber2022editable}
Henrique Weber, Mathieu Garon, and Jean-Fran\c{c}ois Lalonde.
\newblock Editable indoor lighting estimation.
\newblock In {\em ECCV}, 2022.

\bibitem{wiles2020synsin}
Olivia Wiles, Georgia Gkioxari, Richard Szeliski, and Justin Johnson.
\newblock Synsin: End-to-end view synthesis from a single image.
\newblock In {\em CVPR}, 2020.

\bibitem{yu2021luminance}
Hanning Yu, Wentao Liu, Chengjiang Long, Bo Dong, Qin Zou, and Chunxia Xiao.
\newblock Luminance attentive networks for hdr image and panorama
  reconstruction.
\newblock {\em Computer Graphics Forum}, 40(7):181--192, 2021.

\bibitem{yu2021hierarchical}
Piaopiao Yu, Jie Guo, Fan Huang, Cheng Zhou, Hongwei Che, Xiao Ling, and Yanwen
  Guo.
\newblock Hierarchical disentangled representation learning for outdoor
  illumination estimation and editing.
\newblock In {\em ICCV}, 2021.

\bibitem{zhan2021gmlight}
Fangneng Zhan, Yingchen Yu, Rongliang Wu, Changgong Zhang, Shijian Lu, Ling
  Shao, Feiying Ma, and Xuansong Xie.
\newblock Gmlight: Lighting estimation via geometric distribution
  approximation.
\newblock {\em IEEE TIP}, 2022.

\bibitem{zhan2021sparse}
Fangneng Zhan, Changgong Zhang, Wenbo Hu, Shijian Lu, Feiying Ma, Xuansong Xie,
  and Ling Shao.
\newblock Sparse needlets for lighting estimation with spherical transport
  loss.
\newblock In {\em ICCV}, 2021.

\bibitem{zhan2021emlight}
Fangneng Zhan, Changgong Zhang, Yingchen Yu, Yuan Chang, Shijian Lu, Feiying
  Ma, and Xuansong Xie.
\newblock {EMLight: Lighting Estimation via Spherical Distribution
  Approximation}.
\newblock In {\em AAAI}, 2021.

\bibitem{zhang2019fixup}
Hongyi Zhang, Yann~N Dauphin, and Tengyu Ma.
\newblock Fixup initialization: Residual learning without normalization.
\newblock In {\em ICLR}, 2019.

\bibitem{zhang-iccv-17}
Jinsong Zhang and Jean-Fran\c{c}ois Lalonde.
\newblock Learning high dynamic range from outdoor panoramas.
\newblock In {\em ICCV}, 2017.

\bibitem{zhang2019all}
Jinsong Zhang, Kalyan Sunkavalli, Yannick Hold-Geoffroy, Sunil Hadap, Jonathan
  Eisenman, and Jean-Fran{\c{c}}ois Lalonde.
\newblock All-weather deep outdoor lighting estimation.
\newblock In {\em CVPR}, 2019.

\bibitem{zhao2021comodgan}
Shengyu Zhao, Jonathan Cui, Yilun Sheng, Yue Dong, Xiao Liang, Eric~I Chang,
  and Yan Xu.
\newblock Large scale image completion via co-modulated generative adversarial
  networks.
\newblock In {\em ICLR}, 2021.

\bibitem{zhao2020pointar}
Yiqin Zhao and Tian Guo.
\newblock Pointar: Efficient lighting estimation for mobile augmented reality.
\newblock In {\em ECCV}, 2020.

\bibitem{zhu2022irisformer}
Rui Zhu, Zhengqin Li, Janarbek Matai, Fatih Porikli, and Manmohan Chandraker.
\newblock Irisformer: Dense vision transformers for single-image inverse
  rendering in indoor scenes.
\newblock In {\em CVPR}, 2022.

\bibitem{zhu2021spatially}
Yongjie Zhu, Yinda Zhang, Si Li, and Boxin Shi.
\newblock Spatially-varying outdoor lighting estimation from intrinsics.
\newblock In {\em CVPR}, 2021.

\end{thebibliography}
